\pdfoutput=1
\documentclass[10pt,twocolumn,letterpaper]{article}

\usepackage{cvpr}
\usepackage{times}
\usepackage{epsfig}
\usepackage{graphicx}
\usepackage{amsmath}
\usepackage{amssymb}

\usepackage{subfigure}
\usepackage{arydshln}
\usepackage{fmtcount}
\usepackage{algorithm}
\usepackage{algorithmic}
\usepackage{url}
\usepackage{enumitem}
\usepackage{stackengine}
\usepackage{cuted}

\usepackage{amsthm}
\usepackage{color}

\newcommand{\gab}{{G_{H\rightarrow L}}}
\newcommand{\gba}{{G_{L\rightarrow H}}}

\newcommand{\ia}{{I_H}}
\newcommand{\ib}{{I_L}}
\newcommand{\A}{{H}}
\newcommand{\B}{{L}}
\newcommand{\la}{{\mathcal{L}_H}}
\newcommand{\lb}{{\mathcal{L}_L}}
\newcommand{\za}{{Z_H}}
\newcommand{\zb}{{Z_L}}
\newcommand{\ma}{{M_H}}
\newcommand{\mb}{{M_L}}

\newcommand{\fignum}{}
\newcommand{\dirname}{}
\newcommand{\dirnamelarge}{}
\newcommand{\imgwidth}{}
\newcommand{\smallspace}{}
\newcommand{\largespace}{}

\usepackage[breaklinks=true,bookmarks=false]{hyperref}

\cvprfinalcopy 

\begin{document}

\title{Unsupervised Enhancement of Real-World Depth Images Using Tri-Cycle GAN}

\author{Alona~Baruhov\\
Technion -- Israel Institute of Technology\\
{\tt\small alonabar@campus.technion.ac.il}
\and
Guy Gilboa\\
Technion -- Israel Institute of Technology\\
{\tt\small guy.gilboa@ee.technion.ac.il}
}

\maketitle
\begin{abstract}
	Low quality depth poses a considerable challenge to computer vision algorithms. In this work we aim to enhance highly degraded, real-world depth images acquired by a low-cost sensor, for which an analytical noise model is unavailable. In the absence of clean ground-truth, we approach the task as an unsupervised domain-translation between the low-quality sensor domain and a high-quality sensor domain, represented using two unpaired training sets. We employ the highly-successful Cycle-GAN to this task, but find it to perform poorly in this case. Identifying the sources of the failure, we introduce several modifications to the framework, including a larger generator architecture, depth-specific losses that take into account missing pixels, and a novel Tri-Cycle loss which promotes information-preservation while addressing the asymmetry between the domains. We show that the resulting framework dramatically improves over the original Cycle-GAN both visually and quantitatively, extending its applicability to more challenging and asymmetric translation tasks.
\end{abstract}

\section{Introduction}
\label{sec:intro}

\begin{figure}
	\centering
	\renewcommand{\dirname}{./figures/tricycle6/BtoA_example_}
	\renewcommand{\imgwidth}{0.31\linewidth}
	
	\renewcommand{\fignum}{310}
	\subfigure{\includegraphics[width=\imgwidth]{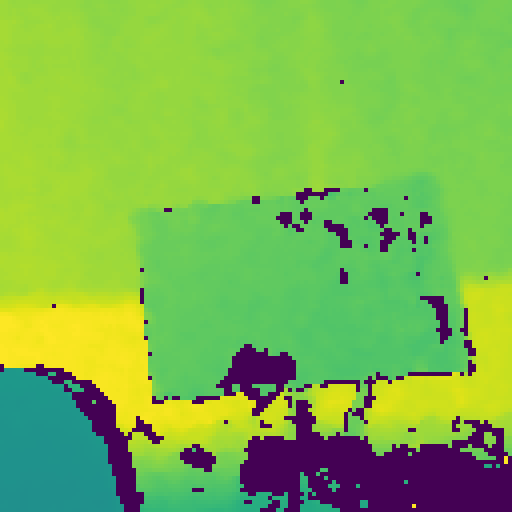}}
	\hfill
	\subfigure{\includegraphics[width=\imgwidth]{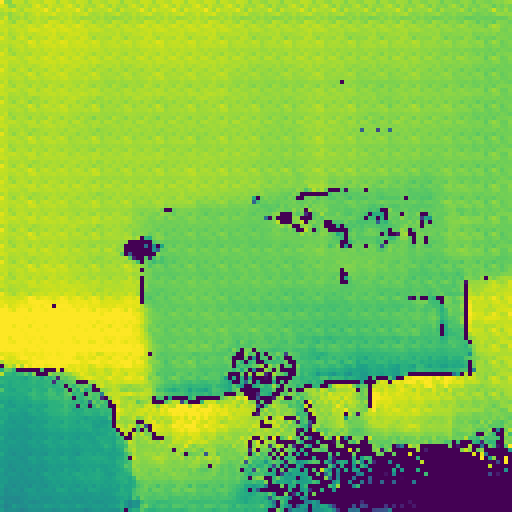}}
	\hfill
	\subfigure{\includegraphics[width=\imgwidth]{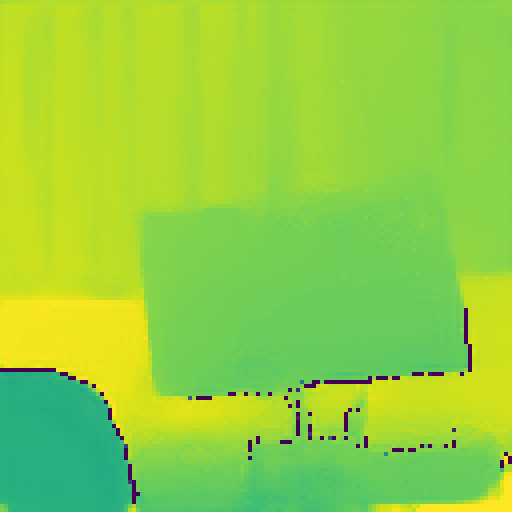}}
	
	\caption{Real-world depth enhancement example. Left to right: \mbox{RealSense} image, original Cycle-GAN result, our result.}
	\label{fig:result-intro}
\end{figure}

Depth information acquired by low-cost depth cameras is typically prone to severe errors and degradations. This low image quality limits the performance of depth-based computer vision algorithms, and challenges most image enhancement methods.
In this work, we aim to enhance these depth images and bring them closer to the output of high-quality depth cameras. We focus on enhancing \emph{real-world} depth images, as produced for example by the Intel Realsense R200 (see Figure~(\ref{fig:result-intro}), left). Due to its small size and low operating power, this camera suffers from substantial noise and artifacts, exhibiting complex and non-random patterns. The absence of any analytic model for these degradations prohibits the use of many classical methods, such as probabilistic and model-based reconstruction methods~\cite{inverseproblems}. Furthermore, it makes simulating realistic degraded depth maps impractical, eliminating the possibility of generating pairs of high- and low-quality images for supervised machine learning algorithms.

As an alternative, we propose a novel approach which eases the requirement for aligned ground-truth image pairs, formulating the task as an \emph{unsupervised} domain-translation problem between a low-quality sensor domain and a high-quality sensor domain. Several works~\cite{deepface, cyclegan, stargan} have recently shown great success in handling such unsupervised domain translation problems. Following their success, we employ a similar approach to the challenging depth enhancement task. We base our approach on the \emph{Cycle-GAN} framework, and develop a fully unsupervised method for training the enhancement network. To the best of our knowledge, this is the first work to formulate this depth enhancement task as an unsupervised translation task.

We focus on the low-power RealSense R200 stereo camera as our low-quality depth sensor. As the high-quality sensor, we select the time-of-flight Microsoft Kinect~2, which is a significantly higher-powered and more accurate camera with substantially less noise. Our aim therefore becomes to bring the quality of the RealSense images to that of the Kinect~2 images via unsupervised domain translation.

Unfortunately, we find the original Cycle-GAN to perform poorly on this task, as depicted in Figure~(\ref{fig:result-intro}) (center). The main sources of this deficiency are the increased complexity of the task, as well as the asymmetry between the domains, manifested by the lack of information equivalence between them. To address these issues, we introduce several modifications to the framework. First, we replace the relatively small generative architecture with a much larger one, with sufficient representational capacity to handle the translation task. Next, we employ depth-specific losses which take into account missing pixels. Finally, we propose the \emph{Tri-Cycle} loss as an alternative information-retention metric for asymmetric domains. Combining these components, our modified Cycle-GAN framework significantly improves over "vanilla" Cycle-GAN in this task, producing much more detailed and less noisy images, as demonstrated in Figure~(\ref{fig:result-intro}) (right). Our main contributions are therefore:
\begin{enumerate}
	\item Developing a training method for depth enhancement networks, capable of handling real-world depth with severe degradation, and without requiring labeled data.
	\item Presenting architectural design principles for CNNs aimed at processing highly degraded depth data with strong non-Gaussian noise, missing pixels, and structured artifacts.
	\item Proposing the \emph{Tri-Cycle loss} that extends the applicability of Cycle-GANs to asymmetric tasks which may not satisfy the information-preservation assumption.

\end{enumerate}

This work is organized as follows. We begin in Section~(\ref{sec:depth-challenges}) with a discussion of the specific challenges of real-world depth and its complex noise sources. We formulate the enhancement problem as an unsupervised translation task in Section~(\ref{sec:unsup-depth-improvement}) and discuss the limitations of the original Cycle-GAN which prevent it from producing reasonable recovery results in this case. We next describe our modifications to the Cycle-GAN framework, including the network architecture and considerations in designing it (Section~(\ref{sec:network-arch})), the depth-specific losses (Section~(\ref{sec:depth-similarity-loss})), and the Tri-Cycle loss (Section~(\ref{sec:tri-cycle})).
As discussed later, the Tri-Cycle loss can be interpreted as a nonlinear generalization of the Moore-Penrose inverse for asymmetrical translation problems, and in our view is the main innovation in this work. We continue by providing experimental results on several datasets in Section~(\ref{sec:results}), demonstrating the effectiveness of the improved Cycle-GAN both visually and quantitatively. We discuss and conclude in Section~(\ref{sec:conclusion}).

\section{Related Work}
\label{sec:related-work}

Depth map completion and enhancement have received considerable attention over the past years. Depth completion methods can generally be divided into two categories: \emph{color-guided} and \emph{non-guided} methods. Color-guided approaches~\cite{deepdepth, sparse2dense, blurrydepth, deeplidar} assume the existence of a color image aligned with the corrupted depth image, and rely on the fact that both share much of the structural information --- such as object edges --- to deduce a dense depth map from the low-quality input. For example,~\cite{deepdepth} uses a CNN to estimate surface normals and edges from the color image, and subsequently combines them with the low quality depth image in a post-process. 
Other works, such as~\cite{sparse2dense}, directly infer the underlying relation between depth and color, and output an enhanced depth map in a single end-to-end process.

When aligned color is not available, either due to a lack of a color sensor, the absence of an alignment between the depth and color streams, low-light conditions, or (as in the RealSense case) the existence of a projected pattern in the visible image, a non-guided completion method must be used~\cite{sparse-depth-sensing, sparse-conv, sparse-dense, sparse-conv-gan}.
For example, Sparse Depth Sensing~\cite{sparse-depth-sensing} reconstructs dense depth maps from very sparse measurements by modeling the scene as a piecewise-planar map, and formulating the recovery task as a compressed sensing problem regularized by sparse second derivatives.

Sparsity-Invariant CNNs~\cite{sparse-conv} take a different approach, and learn an image-to-image enhancement network based on sparse convolutions, which consider only valid depth values when computing convolution outputs.
However, in a follow-up work~\cite{sparse-dense}, the authors note that sparse convolutions rapidly lose their effectiveness after only a few convolutional layers, and thus elect to fill-in missing pixels using a deep architecture based on traditional convolutions instead. The work introduces a unique \emph{sparse training} strategy which synthetically varies the density of valid pixels in the input during training (though in relatively simple patterns), and is found to outperform~\cite{sparse-conv} even at the density for which it was trained.
In parallel, a GAN-based approach was proposed in~\cite{sparse-conv-gan}, introducing an adversarial loss to the supervised depth completion task. The added adversarial loss is shown to notably improve both realism and accuracy of the recovered images compared to previous methods.

Despite the convincing results of all these works in their respective tasks, we note that they all adopt a \emph{supervised} approach to the training process, relying on the availability of ground-truth images alongside the degraded ones. This is often achieved by limiting the method to simple degradations which can be synthetically reproduced, such as i.i.d depth noise and randomly distributed missing pixels. In the case of real-world depth images, however, such assumptions often do not hold. Thus, in this work we take a different approach, and formulate the enhancement task as an \emph{unsupervised} problem which does not require ground-truth images. In this way, we address the task of enhancing depth maps captured by \emph{real world} low-quality depth cameras, and develop a framework for handling this challenging task.

\begin{figure}[t]
	\centering	
	\subfigure{\includegraphics[width=0.4\linewidth]{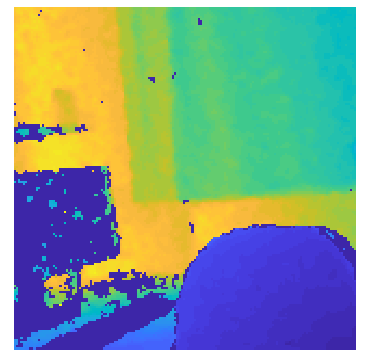}}
	\hspace{0.2in}
	\subfigure{\includegraphics[width=0.4\linewidth]{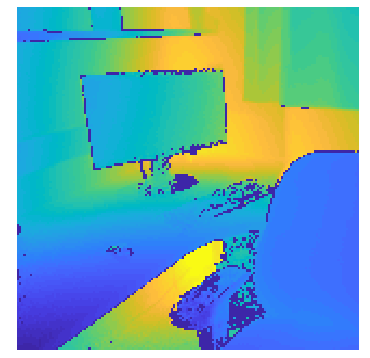}}
	\caption{Left: Example image captured by the RealSense R200 camera. Right: Image of a similar scene captured by the Kinect~2. Images taken from our unpaired \emph{Office} dataset.}
	\label{fig:kinect-rs}
\end{figure}

\section{Improving Depth Images Using Cycle-GAN}

\subsection{The Challenge of Real-World Depth}
\label{sec:depth-challenges}

Low power, small form-factor cameras such as the Intel RealSense R200 suffer from significant noise and artifacts in the captured depth maps. As an active stereo camera, the main sources of error include inaccuracies in the pattern matching --- due to the algorithm itself or to insufficient information in the scene --- as well as from shadowing due to the different viewpoints of the two sensors. These are all amplified by the small camera baseline and the low power of the projector.  
An example image captured with this camera is shown in Figure~(\ref{fig:kinect-rs}) (left).
	
The dependence of the depth noise on multiple factors, including scene-specific details such as texture, material, geometry and lighting, as well as camera-specific parameters such as optics, projector, and algorithm performance, make it virtually impossible to reliably model the depth degradation. Thus, in contrast to many low-level image processing tasks such as denoising or super-resolution, simulating a realistic noisy image given a known ground-truth image is impractical.

In the absence of a viable option to simulate training data, one must resort to manual capturing. One approach could be to capture pairs of images of a scene using two synchronized and calibrated depth cameras, with one being the low-quality camera and the second being a high-quality camera providing the ground truth. Unfortunately, employing such a technique in large scale is extremely complex --- it requires highly accurate alignment of the cameras, suffers from occlusions due to the different viewpoints, and furthermore, since most depth cameras involve some form of active projection, it is impossible to have the two capture the scene at the same time. Consequently, the process becomes lengthy and inefficient, producing too few images to form an effective training set. 
Interestingly, such an aligned dataset was recently presented in~\cite{daniel-kinect-rs}, though to achieve accurate results the process was limited to a specific, highly controlled environment, and resulted in just 112 images. 
	
\subsection{Unsupervised Depth Image Improvement}
\label{sec:unsup-depth-improvement}

Considering the huge challenge in producing pairs of input-output images for real-world depth enhancement, we believe that the most viable path for training such a process is \emph{unsupervised} learning. In this approach, the problem is re-stated as a translation problem between two domains --- a low-quality domain $L$ and a high-quality domain $H$, represented by two \emph{unaligned}, freely captured training sets. Such translation tasks have recently received significant attention, and have shown remarkable results in many translation problems~\cite{cyclegan,discogan,dualgan,stargan,s-flowgan}.

Following previous work, we adopt the highly successful Cycle-GAN~\cite{cyclegan,discogan,dualgan} as the basis for our domain translation framework. The Cycle-GAN simultaneously learns \emph{two} generative networks for translating in both directions, and uses cycle-consistency to encourage information-preservation by the translation in the absence of ground-truth targets. Specifically, given the two domains $L$ and $H$, the loss function of the Cycle-GAN is given by:
\begin{align}
\mathcal{L}(&\gba,~\gab) = \nonumber \\
& \qquad\lambda_1 \la(\gba) + \lambda_2 \lb(\gab) + \nonumber \\
& \qquad\lambda_3 \mathbb{E}_{\ia\in {\A}} \|\gba(\gab(\ia)) - \ia\| + \nonumber \\
& \qquad\lambda_4 \mathbb{E}_{\ib\in {\B}} \|\gab(\gba(\ib)) - \ib\| + \nonumber \\
& \qquad\lambda_5 \mathbb{E}_{\ia\in {\A}} \|\gba(\ia)-\ia\| + \nonumber \\
& \qquad\lambda_6 \mathbb{E}_{\ib\in {\B}} \|\gab(\ib)-\ib\| ~.
\label{eq:cycle-gan}
\end{align}
Here, $\gba$ and $\gab$ are the two learned translators, $\ib$ and $\ia$ are images from the two domains, and $\lb$ and $\la$ are adversarial losses~\cite{gan} for their respective domains, each incorporating a learned discriminator working against the generator (we omit the full definition of this loss for conciseness). The first two losses in this formulation guide the translators to output images in their correct domains (represented via the exemplar images from each domain), the next two losses are the cycle-consistency losses, and the final two losses are the identity losses which regularize the training process, and were introduced in~\cite{cyclegan}.

The nature of the depth data, however, poses significant challenges to the Cycle-GAN framework. First, the low-quality depth exhibits significantly stronger and more complex noise patterns than traditionally used with the Cycle-GAN, and large missing regions create severe discontinuities in the data. Furthermore, we observe that the information preservation assumption made by the Cycle-GAN design does not actually hold in our case --- specifically, the two depth domains are in fact \emph{not} equivalent, with the high-quality domain containing distinctly more information than the low-quality one. Thus, the cycle-consistency constraint which forms the basis of the Cycle-GAN becomes problematic in this case.
To address these issues, we modify several key aspects of the original Cycle-GAN formulation, enabling it to successfully handle this challenging task. In the next sections we detail these modifications.

\subsubsection{Network Architecture}
\label{sec:network-arch}

Since a large part of the difficulty in low-quality depth comes from the high number of missing pixels, one may be tempted to consider architectures based on sparse convolutions~\cite{sparse-conv,partialconv}, which are a type of layer specifically designed for inpainting problems. In a masked convolution, only known pixels contribute to the result of the convolution, with each output feature normalized by the number of contributing values. However, for our task of depth enhancement, we have found such networks to perform poorly. Figure~(\ref{fig:kinect-rs}) (left) reveals a possible explanation: as opposed to inpainting tasks where the hole locations are typically arbitrary, in depth images the holes are in fact \emph{strongly correlated} with the properties and geometry of the scene. In other words, the holes themselves convey information about the objects being recovered, such as their shape or distance. Thus, masking this information using convolutions invariant to the hole configuration is actually counter-productive in our case, and does not contribute to the desired result.

\begin{figure}
	\centering
	\includegraphics[width=0.83\linewidth]{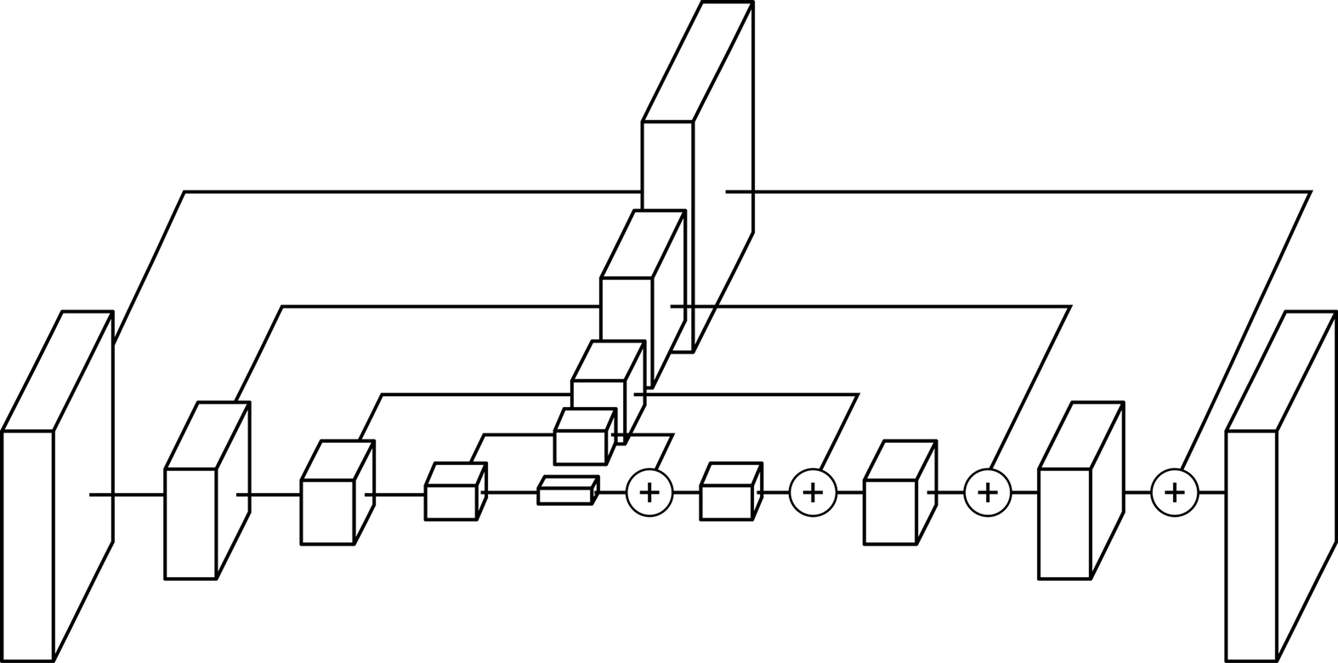}
	\caption{U-Net architecture with skip connections (schematic).}
	\label{fig:unet2}
\end{figure}

With this understanding, we base our translation network on standard convolutions, and consider the entire depth image --- including its structured zero values --- as a single visual representation of the scene. We employ a standard U-Net with skip connections~\cite{unet,hourglass} as the generator architecture, similar to the original Cycle-GAN. The basic U-Net architecture is illustrated in Figure~(\ref{fig:unet2}).

However, plugging-in a simple U-Net to the Cycle-GAN produces strikingly bad results in our case. To handle the complexity of low-quality depth, it is crucial to use a much wider and deeper translation network. Specifically, we significantly increase the number of channels in the lower layers of the network --- those that respond to high frequencies in the image --- to enable the network to more effectively handle the large variety of local patterns that emerge in the presence of holes. At the same time, we use a much deeper architecture than typically used in Cycle-GANs to allow the network to better resolve large object-scale phenomena, which is required to reliably fill-in large holes and compensate for complex artifacts. 
Our full generator architecture is detailed in Table~(\ref{tbl:unet-advanced}).

\begin{table}[!t]
	\renewcommand\arraystretch{1}
	\setlength{\tabcolsep}{8pt}
	\centering
	\footnotesize
	\begin{tabular}{|c|c|c|}
		\hline
		\textbf{Layer Name} & \textbf{Input Layers} & \textbf{Output Size} \\
		\hline
		input & - & $128\times 128\times 1$ \\
		\hline
		conv1 & input & $128\times 128\times 128$ \\
		\hline
		conv2.1 & conv1 & $64\times 64\times 128$ \\
		\hline
		conv2.2 & conv2.1 & $64\times 64\times 128$ \\
		\hline
		conv3.1 & conv2.2 & $32\times 32\times 128$ \\
		\hline
		conv3.2 & conv3.1 & $32\times 32\times 128$ \\
		\hline
		conv4.1 & conv3.2 &  $16\times 16\times 128$ \\
		\hline
		conv4.2 & conv4.1 &  $16\times 16\times 128$ \\
		\hline
		conv5.1 & conv4.2 & $8\times 8\times 128$ \\
		\hline
		conv5.2 & conv5.1 & $8\times 8\times 128$ \\
		\hline
		conv6.1 & conv5.2 & $4\times 4\times 128$ \\
		\hline
		conv6.2 & conv6.1 & $4\times 4\times 128$ \\
		\hline
		conv7.1 & conv6.2 & $2\times 2\times 128$ \\
		\hline
		conv7.2 & conv7.1 & $2\times 2\times 128$ \\
		\hline
		conv6.3 & up(conv7.2) $+$ conv(conv6.2) & $4\times 4\times 128$ \\
		\hline
		conv6.4 & conv6.3 & $4\times 4\times 128$ \\
		\hline
		conv5.3 & up(conv6.4) $+$ conv(conv5.2) & $8\times 8\times 128$ \\
		\hline
		conv5.4 & conv5.3 & $8\times 8\times 128$ \\
		\hline
		conv4.3 & up(conv5.4) $+$ conv(conv4.2) &  $16\times 16\times 128$ \\
		\hline
		conv4.4 & conv4.3 & $16\times 16\times 128$ \\
		\hline
		conv3.3 & up(conv4.4) $+$ conv(conv3.2) & $32\times 32\times 128$ \\
		\hline
		conv3.4 & conv3.3 & $32\times 32\times 128$ \\
		\hline
		conv2.3 & up(conv3.4) $+$ conv(conv2.2) & $64\times 64\times 128$ \\
		\hline
		conv2.4 & conv2.3 & $64\times 64\times 128$ \\
		\hline
		conv1.3 & up(conv2.4) $+$ conv(conv1.2) & $128\times 128\times 128$ \\
		\hline
		conv1.4 & conv1.3 & $128\times 128\times 1$ \\
		\hline
	\end{tabular}
	\caption{Generator architecture. Each layer consists of a $3\times 3$ convolution, Leaky ReLU, and instance normalization~\cite{instancenorm}, with a stride of 1 or 2 depending on the output size. The $+$ operator represents channel-wise concatenation, up$(\cdot)$ denotes $\times 2$ nearest-neighbor upsampling, and conv$(\cdot)$ denotes a size-maintaining convolution with Leaky ReLU and instance normalization.}
	\label{tbl:unet-advanced}
\end{table}

\subsubsection{Depth-Specific Losses}

\label{sec:depth-similarity-loss}

\begin{figure}
	\centering
	\includegraphics[width=0.4\linewidth]{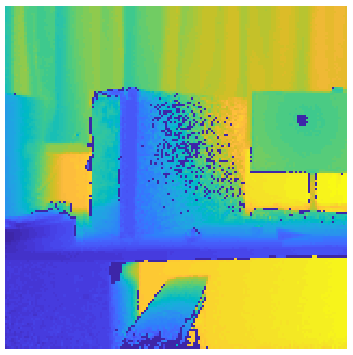}
	\caption{Kinect~2 image with scattered holes (in dark blue).}
	\label{fig:kinect-noise}
\end{figure}

The Cycle-GAN uses image similarity as a central component in the training process, in both the cycle-consistency and identity losses. However, when missing pixels are involved, computing similarity over the entire image may be suboptimal, particularly for pixels which are scattered and random. We note that while so far we have focused mainly on the structured noise of the \mbox{RealSense} camera, the Kinect~2 camera suffers from noise as well. Specifically, while the Kinect images exhibit significantly fewer holes than the RealSense images, and though some of these holes follow object boundaries and discontinuities, many of them are random and isolated, as demonstrated in Figure~(\ref{fig:kinect-noise}). These random patterns are due to the time-of-flight technology, which often forms holes in areas of low reflectivity, or where external light sources overpower the camera's own projector. Clearly, requiring a generator to re-create these precise random patterns, for instance in the Kinect~$\rightarrow$~RealSense~$\rightarrow$ Kinect cycle, would be counter-productive as it would force the first translator to encode "hints" about the original hole locations in the RealSense image.

To address this, we utilize \emph{masked similarity}, which considers only non-zero locations when computing distance. 
Formally, given a known depth image $Z$ with valid pixel mask~$M:=\big(z_{i,j} \neq 0\big)_{i,j}$, and given a second depth image $\tilde{Z}=f(Z)$, we define the masked similarity loss as
\begin{equation}
d(Z,\tilde{Z};M)=\|M\circ(Z-\tilde{Z})\|_1~,
\label{eq:valid-loss}
\end{equation}
where $\circ$ denotes element-wise (Hadamard) multiplication.

We note that~(\ref{eq:valid-loss}) is not symmetric in $Z$ and $\tilde{Z}$. We use a non-symmetric loss since the valid pixel mask $\tilde{M}=M(\tilde{Z})$ of the output depth is non-differentiable in the network parameters and is unstable near $0$, and thus optimizing with respect to it would be impractical.  
Furthermore, the symmetric variant would encourage the formation of holes in the output, and in fact has a trivial global minimum at $\tilde{Z}=0$. In contrast, the asymmetric similarity generally prefers filling-in holes in the output image, while still allowing isolated holes to form owing to the robust $L_1$ norm. 

Finally, an additional issue we have observed with the original Cycle-GAN is \emph{range preservation}. Specifically, for any solution $(\gab,~\gba)$ of~(\ref{eq:cycle-gan}), the solution $(\gab+z_0,~\gba-z_0)$, where $\pm z_0$ represents a depth shift by~$z_0$ of the non-zero values, is also equally valid.
To counter this effect, we add a small masked similarity loss to the $L\rightarrow H$ translation, requiring that the high-quality image be close to the low-quality one where it is non-zero. Formally, this loss is given by:
\begin{equation}
\mathbb{E}_{\zb\in {\B}} \|\mb\circ(\gba(\zb)-\zb)\|_1~,
\end{equation}
where $\mb$ is the valid mask of the low-quality image. We note that since this image typically has significantly more holes than the expected high-quality output, this loss essentially just maintains the overall distance of the known objects in the scene, without affecting the visual properties of the output image.

\subsubsection{Tri-Cycle Loss}
\label{sec:tri-cycle}

\begin{figure}
	\centering
	\includegraphics[width=\linewidth]{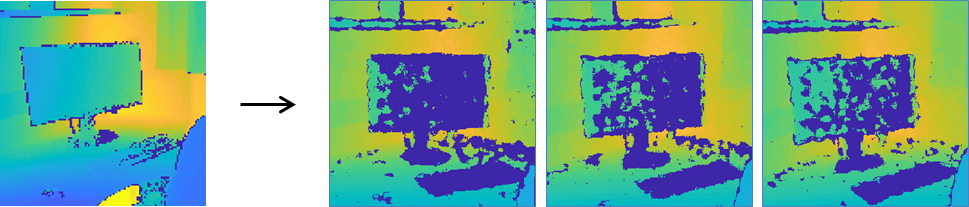}
	\caption{Kinect-to-RealSense as a \emph{one-to-many} mapping.}
	\label{fig:one2many}
\end{figure}

The Cycle-GAN measures information preservation by passing images through a full cycle of the domain translation process, and requiring the result to be an identity operator. However, particularly in the $H\rightarrow L$ translation, this transform is in fact a \emph{one-to-many} mapping, as the low quality image may degrade in many different ways. An example of this is shown in Figure~(\ref{fig:one2many}). Formally, for a 3D scene $S$ and viewpoint $V$, the depth image is ideally a projection \mbox{$Z=P(S;V)$} of the scene on the camera plane. However, due to noise and errors in the capturing process, we obtain a measured depth image which can very roughly be expressed as \mbox{$\hat{Z} = M\circ(Z+n)$}, with~$n$ the depth noise and $M$ a mask image. Thus, for a fixed scene and viewpoint, we may measure any one of \emph{many} possible depth images $\{\hat{Z}\}$. In the presence of holes and strong depth errors, this set can become of significant size, in contrast to e.g., a color camera where the noise model can be approximated as a Gaussian or Poisson source with typically low variance, leading to a relatively compact set.

To address this, we propose using an \emph{asymmetric} loss function for promoting information preservation, which does not require the two translations to be inverses. Instead, this loss essentially requires that when performing a full cycle of the form $L~\rightarrow~H~\rightarrow~L$, we simply produce a low-quality image which could \emph{feasibly} reproduce the high-quality one, but not necessarily the same one we began with.

To this end, we regard the Kinect camera as a high-quality camera with relatively low noise, and hence consider the volume of the set $\{\hat{Z}\}$ to be negligible. However, this does not hold for the low quality RealSense camera, where the capturing process is substantially less stable, and multiple frames of the same scene may display large variation.

\begin{table*}[h!]
	\begin{align}
	\mathcal{L}(& \gba, ~\gab) = & \nonumber \\
	& ~\quad\lambda_1 \la(\gba) + \lambda_2 \lb(\gab) + & \text{\emph{(Adversarial)}} \nonumber \\
	& ~\quad\lambda_3 \mathbb{E}_{\za\in {\A}} \|\ma\circ(\gba(\gab(\za))-\za)\|_1 + & \text{\emph{(H cycle)}} \nonumber \\
	& ~\quad\lambda_4 \mathbb{E}_{\zb\in {\B}} \|\gab(\gba(\zb))-\zb\|_1 + & \text{\emph{(L cycle)}} \nonumber \\
	& ~\quad\lambda_5 \mathbb{E}_{\za\in {\A}} \|\ma\circ(\gba(\za)-\za)\|_1 + & \text{\emph{(H identity)}} \nonumber \\
	& ~\quad\lambda_6 \mathbb{E}_{\zb\in {\B}} \|\gab(\zb)-\zb\|_1 ~+ & \text{\emph{(L identity)}} \nonumber \\
	& ~\quad\lambda_7 \mathbb{E}_{\zb\in {\B}} \|\mb\circ(\gba(\zb)-\zb)\|_1 & \text{\emph{(Depth preserve)}} \nonumber \\
	& ~\quad\lambda_8 \mathbb{E}_{\zb\in {\B}} \|\gba(\gab(\gba(\zb))) - \gba(\zb)\|_1 & \text{\emph{(Tri-cycle)}} \nonumber
	\end{align}
	\caption{Full loss function for the depth enhancement Tri-Cycle GAN.}
	\label{fig:full-loss}
\end{table*}

Given a low-quality depth image $Z$ captured from the underlying scene $S=S(Z)$ and viewpoint $V=V(Z)$, we denote by $\psi(Z)$ the set of \emph{all} low quality images which could have been captured under the same conditions:
\begin{equation}
\psi(Z) = \Big\{ \hat{Z} = M\circ(P(S;V) + n) ~~ \big| ~~ (M,n) \sim \mathcal{P}_L(S,V) \Big\}~.
\end{equation}
Here, $\mathcal{P}_L(S,V)$ denotes the joint distribution of plausible low-quality depth noise and hole patterns corresponding to the underlying scene. The set $\psi(Z)$ forms the \emph{equivalency set} of $Z$, and as previously noted, has a non-negligible volume due to the properties of the low-quality depth camera.

Returning to the Cycle-GAN formulation~(\ref{eq:cycle-gan}), it is now evident that the cycle-consistency assumption is broken in the $L~\rightarrow~H~\rightarrow~L$ case, as the second translation is a one-to-many mapping. Clearly, requiring this cycle to be the identity mapping is an unnecessarily difficult constraint. 
We thus propose to \emph{relax} the cycle-consistency constraint in this case, such that the output \emph{belongs to the equivalency set} of the input, rather than equal it. This translates to the constraint:
\begin{equation}
\gab(\gba(\zb))\in\psi(\zb)~.
\end{equation}
Unfortunately, enforcing this constraint directly is impractical, as the set $\psi(\zb)$ is a complex, non-convex set with no analytical form.
However, if we apply $\gab$ to both sides of this expression, and by using the fact that $\gba(\hat{Z}_L)=\gba(\zb)$ iff $\hat{Z}_L\in\psi(\zb)$, we can re-write the above as:
\begin{equation}
\gba(\gab(\gba(\zb))) = \gba(\zb)~.
\label{eq:tri-cycle-constraint}
\end{equation}
This requirement readily translates to a loss function, which we term the \emph{Tri-Cycle loss} due to the application of three consecutive translations in its definition. Incorporating an $L_1$ distance norm, and accumulating over the entire low-quality domain, this loss becomes:
\begin{align}
&\mathcal{L}_{tri}(\gba, \gab) = \\
&\quad\mathbb{E}_{\zb\in {\B}} \left\|\gba(\gab(\gba(\zb))) - \gba(\zb)\right\|_1. \nonumber
\end{align}

It is interesting to note the similarity between the above Tri-Cycle loss and the linear Generalized Inverse. In the linear case, we may consider the inversion of a dimension-reducing matrix (i.e., a many-to-one mapping) \mbox{$A: x\in\mathbb{R}^n \rightarrow y\in\mathbb{R}^m$} with $m<n$, with the set $\psi(x)$ in this case being the Affine set of all vectors $x$ mapped to the same $y$. The Penrose conditions for inverting such a matrix~\cite{generalized-inverse} essentially require that the inverse transform map every $y=Ax$ to \emph{one} of the vectors which would have been mapped to it by $A$, i.e., $A^+y=A^+Ax \in \psi(x)$, which is formalized by the condition $AA^+Ax=Ax$. Indeed, our Tri-Cycle constraint follows very similar reasoning. 
In this sense, we may view the mapping $\gab$ as a \emph{generalized inverse} of $\gba$, and the Tri-Cycle formulation as seeking one of these mappings as part of the optimization process.

\subsection{Full Loss Function and Optimization Method}
\label{sec:final-loss}

Our full depth enhancement architecture optimizes a combined penalty consisting of all the losses discussed in the previous sections. The full loss function is given in Table~(\ref{fig:full-loss}). We optimize using ADAM~\cite{adam} with batch size~1, with each batch consisting of a pair of random low-quality and high-quality images sampled from $\B\times \A$. 
We use a constant learning rate of~0.001, and augment the examples with random crops, 90-degree rotations, horizontal and vertical flips, and random shifts in depth.

\section{Experimental Evaluation and Results}
\label{sec:results}

\subsection{Synthetic Experiments}
\label{sec:synthetic-res}

\begin{figure}
	\centering
	\renewcommand{\largespace}{-0.08in}
	
	\subfigure{\includegraphics[width=0.23\linewidth]{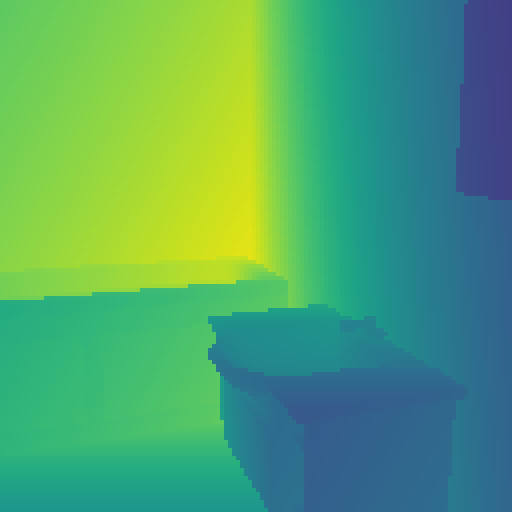}}
	\hfill
	\subfigure{\includegraphics[width=0.23\linewidth]{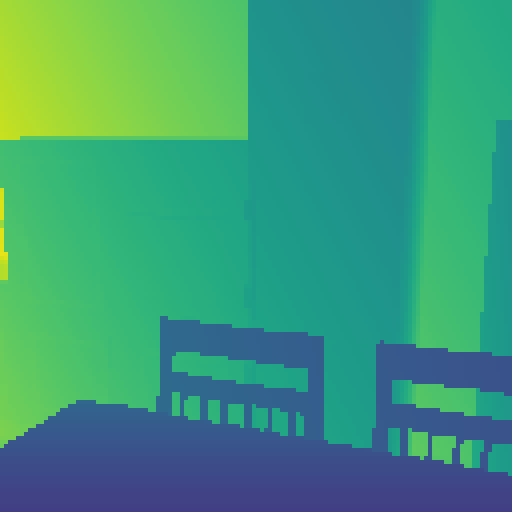}}
	\hfill
	\subfigure{\includegraphics[width=0.23\linewidth]{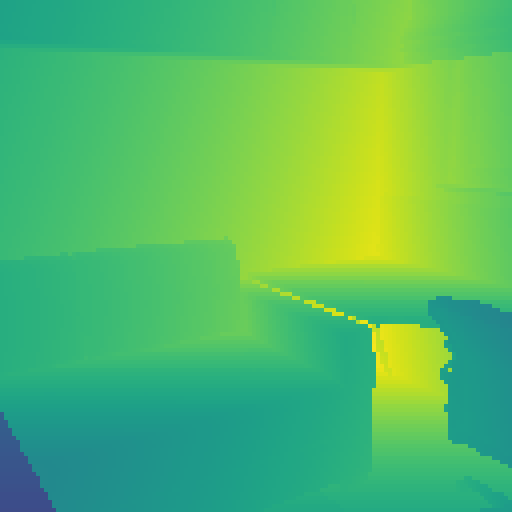}}
	\hfill
	\subfigure{\includegraphics[width=0.23\linewidth]{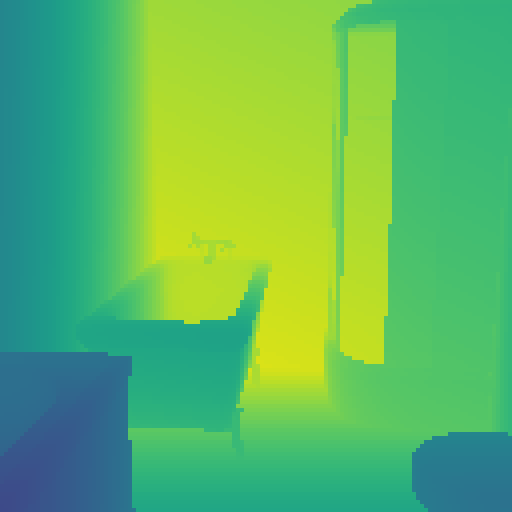}}
	
	\vspace{\largespace}
	
	\subfigure{\includegraphics[width=0.23\linewidth]{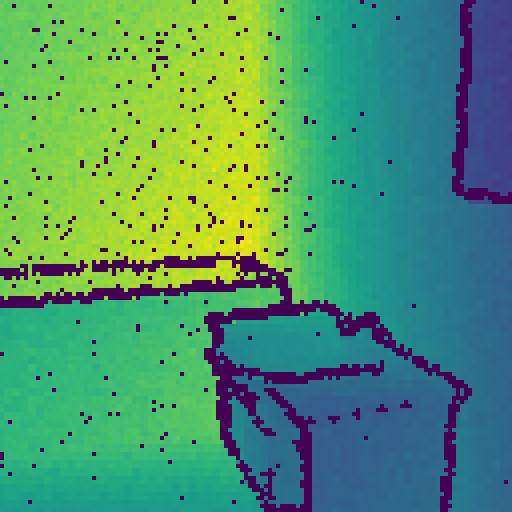}}
	\hfill
	\subfigure{\includegraphics[width=0.23\linewidth]{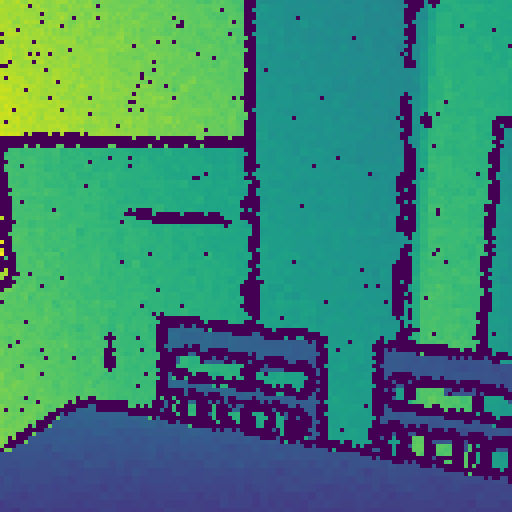}}
	\hfill
	\subfigure{\includegraphics[width=0.23\linewidth]{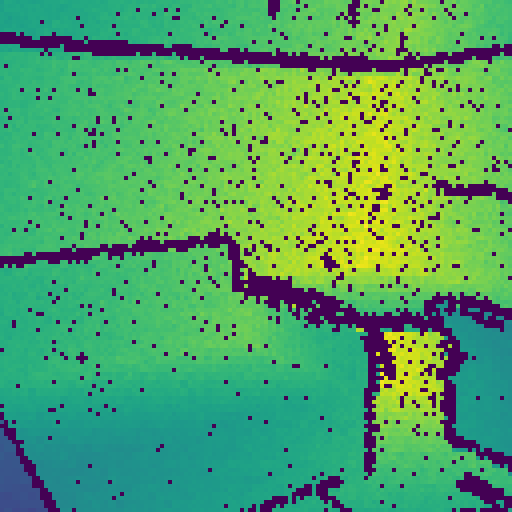}}
	\hfill
	\subfigure{\includegraphics[width=0.23\linewidth]{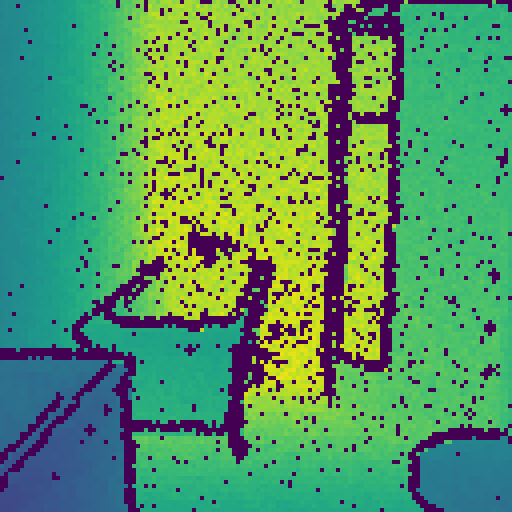}}
	
	\caption{Some examples from our synthetic dataset. Top: Original rendered images. Bottom: Images with synthetically added depth noise.}
	\label{fig:synthetic-data}
\end{figure}

To quantify the performance of our Cycle-GAN framework, we use a high-quality dataset of rendered depth images, to which we apply noise in a process simulating a depth camera. 
Our dataset is based on the Physically Based Rendering Dataset~\cite{suncg-rendered-website} consisting of 568,793 depth images randomly sampled from the SUNCG set of 3D scenes~\cite{suncg-website}. We further filter the data by removing images with very low standard deviation ($\sigma<$ 400mm) or with more than 15\% distant pixels ($d>$ 5000mm), as we are emulating a depth camera with limited range. The resulting synthetic dataset contains around 120,000 images, see Figure~(\ref{fig:synthetic-data}) (top).

For the depth noise, we apply several degradations typical of depth cameras. Unfortunately, it is extremely difficult to emulate the highly structured noise of the RealSense camera. However, our process includes several noise sources which are common to depth cameras such as the RealSense and Kinect. These include \emph{structural noise}, generated by adding Gaussian noise to a down-sampled version of the image, followed by nearest-neighbor upsampling; \emph{object boundary noise}, produced by removing pixels near object edges with a probability of $p=80\%$; \emph{depth-adaptive noise}, generated as random Gaussian noise with a distance-dependent standard deviation $\sigma \sim z^3$; and \emph{depth-adaptive holes}, generated by randomly eliminating pixels from the image with a probability $p\sim z^4$.
Figure~\ref{fig:synthetic-data} (bottom) shows a few noisy images produced by this process.

We train a translation network to convert between the noisy and noiseless depth domains. Our experiments compare the performance of our full Cycle-GAN framework to the original formulation, as well as to the original Cycle-GAN but with the larger generator architecture. 
In addition, we compare our results to those of the recent Sparse Depth Sensing depth enhancement algorithm~\cite{sparse-depth-sensing}.

For the quantitative comparison, it is well-known that traditional measures such as PSNR are unreliable as image quality estimators, particularly when adversarial and perceptual losses are involved~\cite{blau-rethinking,superres,ct,compression}. Indeed, in most cases methods which directly optimize MSE will outperform perceptual methods in terms of PSNR, while in reality they produce over-smoothed images which lack detail. Thus, to more accurately quantify recovery of detail, we instead propose a \emph{patch-based normalized cross-correlation} (PNCC) measure. This measure computes the similarity between two images by computing the normalized cross-correlations between their local patches (with overlap), and averaging the results. Formally, given two images $X$ and $Y$, we define $PNCC(X,Y)$ in terms of a block size~$b$ and a step size~$s$.
Denoting by $I^b_{i,j}$ the patch of image $I$ beginning at $(i,j)$ and extending to $(i+b-1,j+b-1)$ (inclusive), the similarity between $X$ and $Y$ is computed as:
\begin{align}
PNCC(X,Y) = \frac{1}{N}\sum_{\begin{smallmatrix}
	i\in [0, s, 2s, \ldots] \\ 
	j\in [0, s, 2s, \ldots]
	\end{smallmatrix}}
\rho\left(X^b_{i,j}, Y^b_{i,j}\right)
\end{align}
Here, $\rho(\cdot,\cdot)$ is the normalized cross correlation function and $N$ is the total number of patches in the sum.

\begin{figure}
	\centering
	\renewcommand{\dirname}{./figures/synthetic-comparison-0.1-scaled-4/BtoA_example_}
	\renewcommand{\dirnamelarge}{./figures/synthetic-comparison-0.1-scaled-4-enlargements/BtoA_example_}
	\renewcommand{\smallspace}{-0.08in}
	\renewcommand{\largespace}{-0.01in}
	\renewcommand{\imgwidth}{0.19\linewidth}
	
	\renewcommand{\fignum}{121}
	\subfigure{\includegraphics[width=\imgwidth]{\dirname\fignum/img_0.png}}
	\hfill
	\subfigure{\includegraphics[width=\imgwidth]{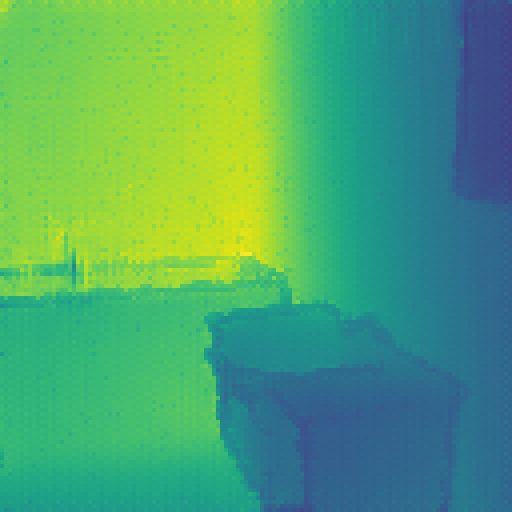}}
	\hfill
	\subfigure{\includegraphics[width=\imgwidth]{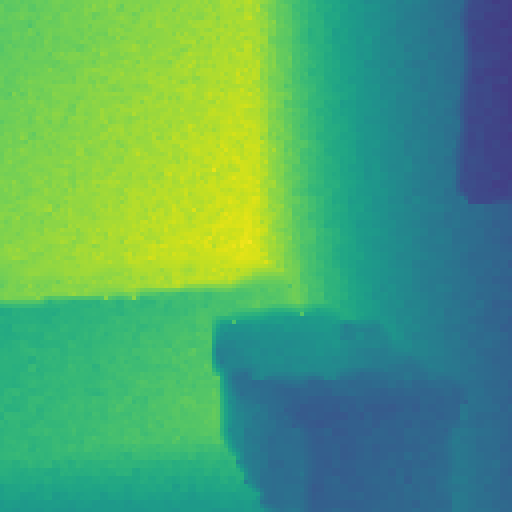}}
	\hfill
	\subfigure{\includegraphics[width=\imgwidth]{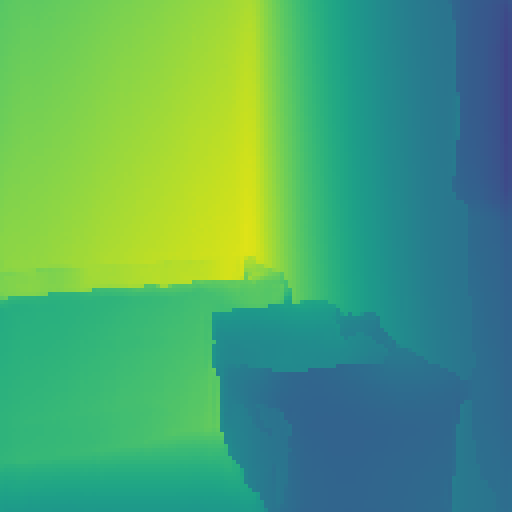}}
	\hfill
	\subfigure{\includegraphics[width=\imgwidth]{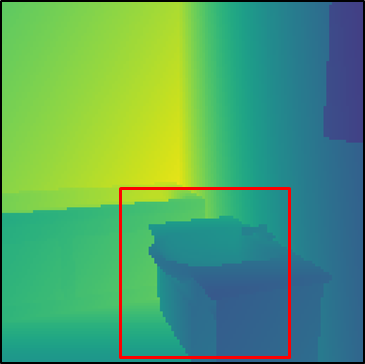}}
	
	\vspace{\smallspace}
	
	\subfigure{\includegraphics[width=\imgwidth]{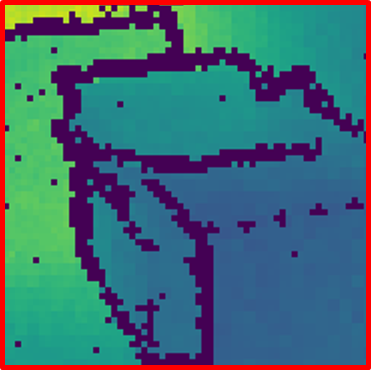}}
	\hfill
	\subfigure{\includegraphics[width=\imgwidth]{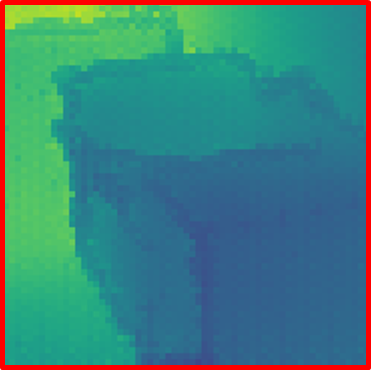}}
	\hfill
	\subfigure{\includegraphics[width=\imgwidth]{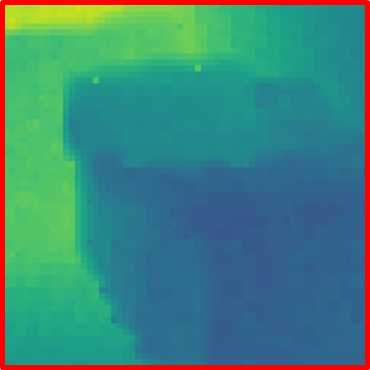}}
	\hfill
	\subfigure{\includegraphics[width=\imgwidth]{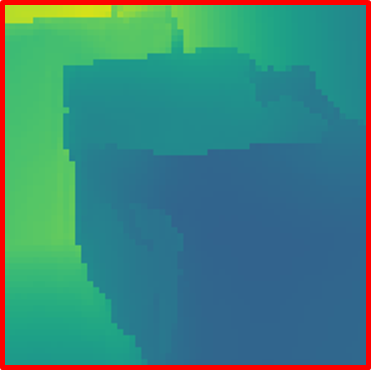}}
	\hfill
	\subfigure{\includegraphics[width=\imgwidth]{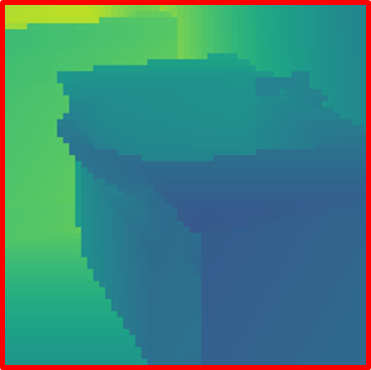}}
	
	\vspace{\largespace}
	
	\renewcommand{\fignum}{976}
	\subfigure{\includegraphics[width=\imgwidth]{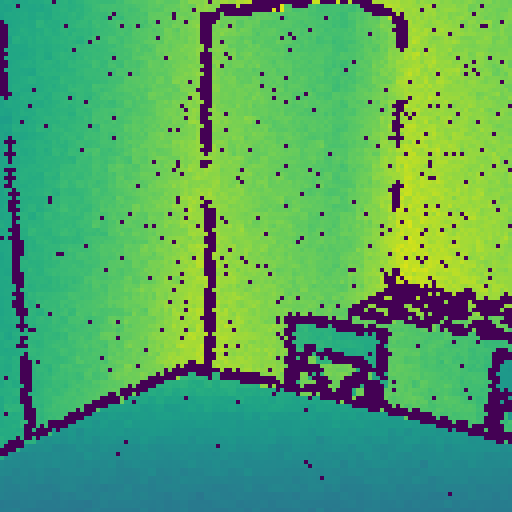}}
	\hfill
	\subfigure{\includegraphics[width=\imgwidth]{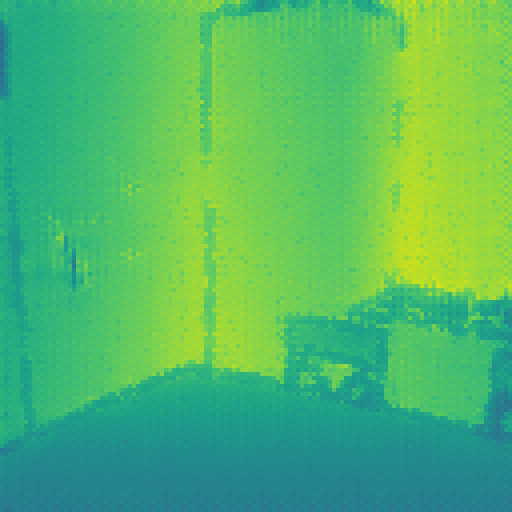}}
	\hfill
	\subfigure{\includegraphics[width=\imgwidth]{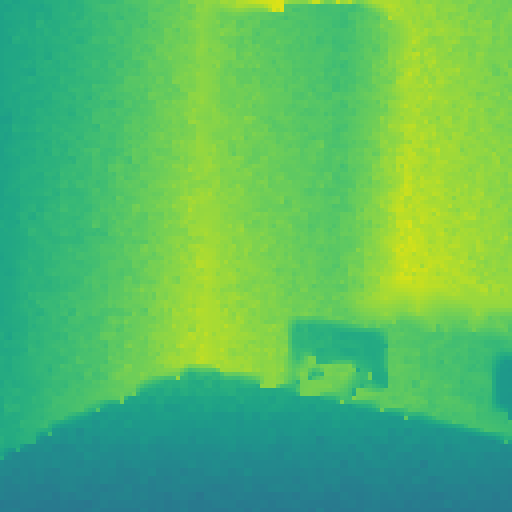}}
	\hfill
	\subfigure{\includegraphics[width=\imgwidth]{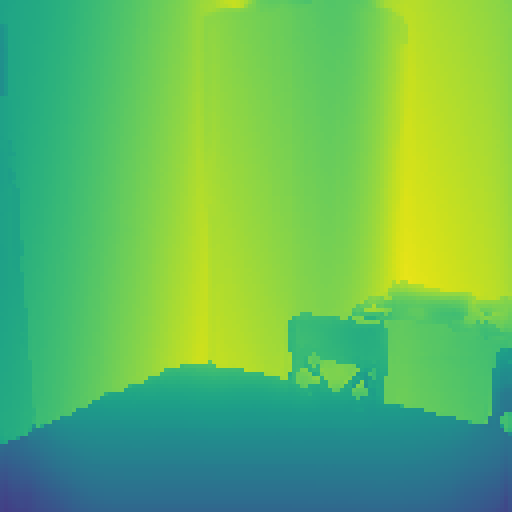}}
	\hfill
	\subfigure{\includegraphics[width=\imgwidth]{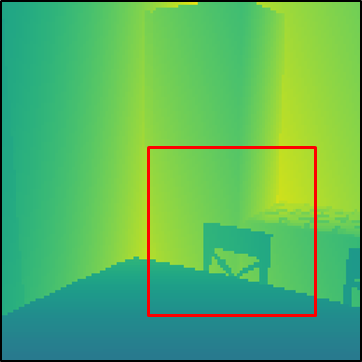}}
	
	\vspace{\smallspace}
	
	\subfigure{\includegraphics[width=\imgwidth]{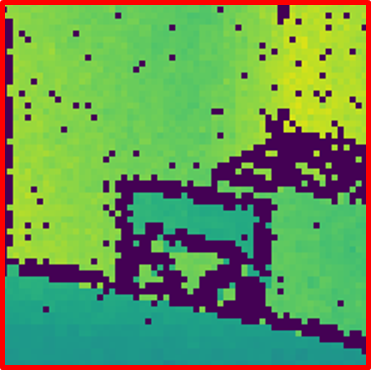}}
	\hfill
	\subfigure{\includegraphics[width=\imgwidth]{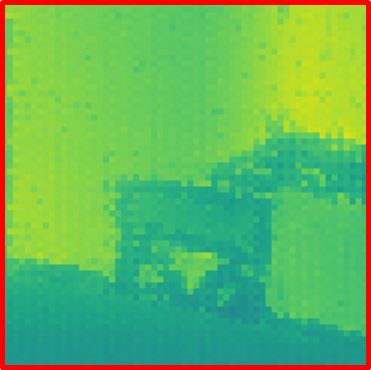}}
	\hfill
	\subfigure{\includegraphics[width=\imgwidth]{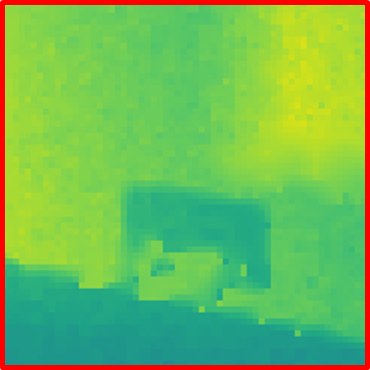}}
	\hfill
	\subfigure{\includegraphics[width=\imgwidth]{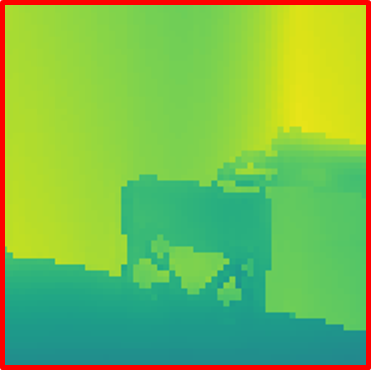}}
	\hfill
	\subfigure{\includegraphics[width=\imgwidth]{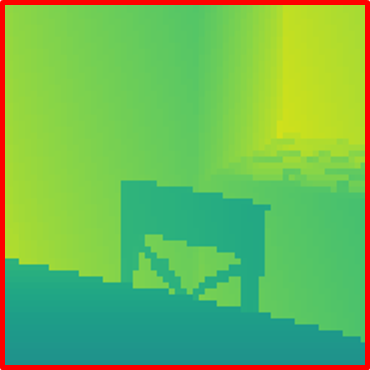}}
	
	\vspace{\largespace}

	\renewcommand{\fignum}{2533}
	\subfigure{\includegraphics[width=\imgwidth]{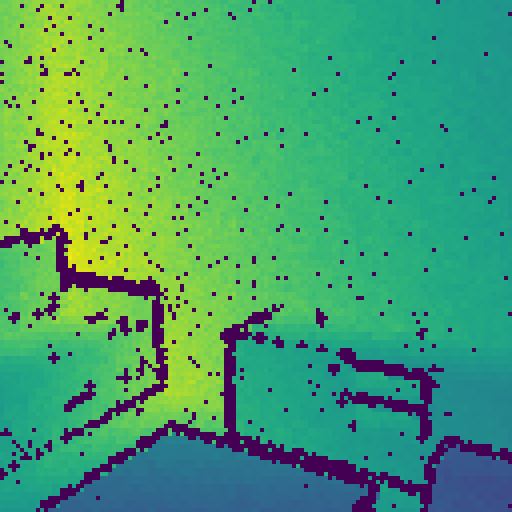}}
	\hfill
	\subfigure{\includegraphics[width=\imgwidth]{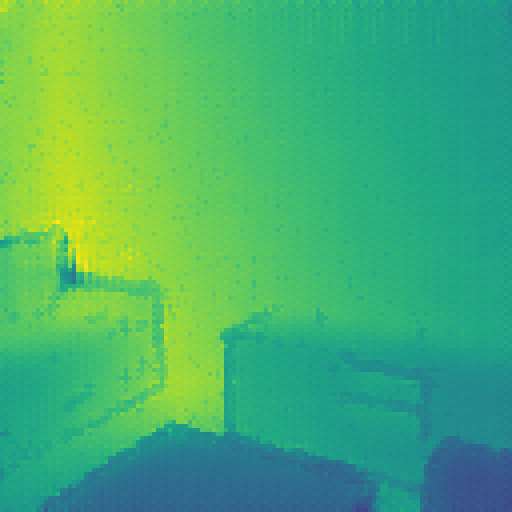}}
	\hfill
	\subfigure{\includegraphics[width=\imgwidth]{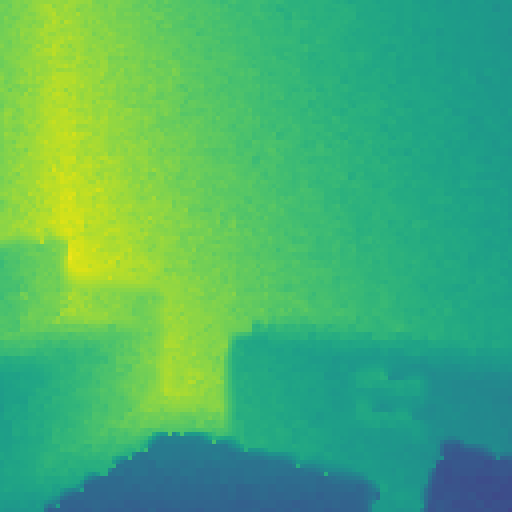}}
	\hfill
	\subfigure{\includegraphics[width=\imgwidth]{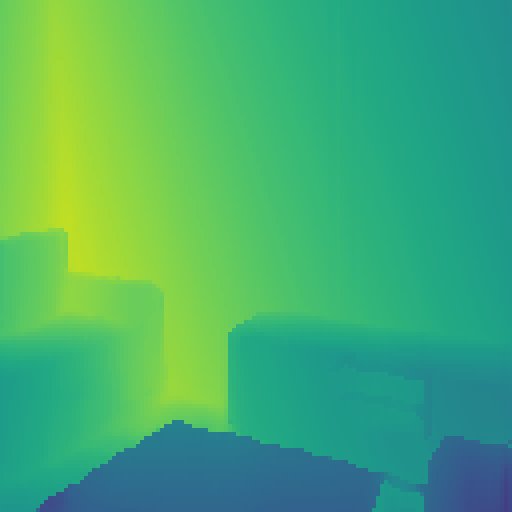}}
	\hfill
	\subfigure{\includegraphics[width=\imgwidth]{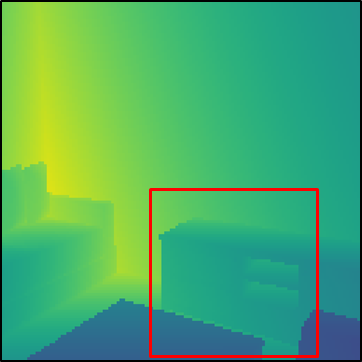}}	
	
	\vspace{\smallspace}
	
	\subfigure{\includegraphics[width=\imgwidth]{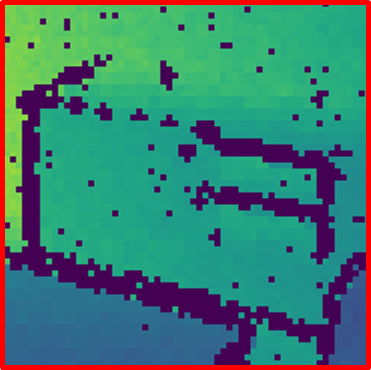}}
	\hfill
	\subfigure{\includegraphics[width=\imgwidth]{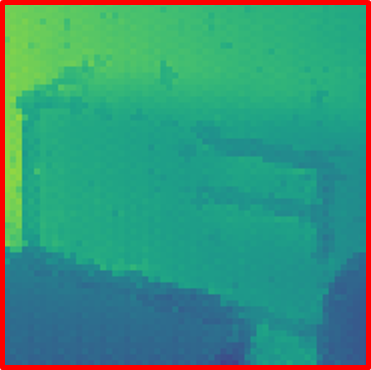}}
	\hfill
	\subfigure{\includegraphics[width=\imgwidth]{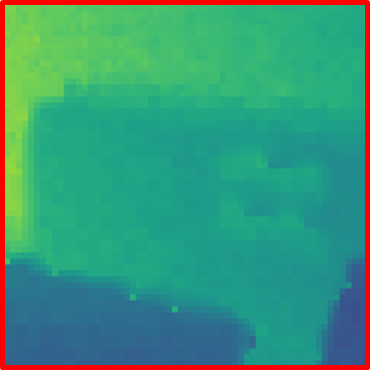}}
	\hfill
	\subfigure{\includegraphics[width=\imgwidth]{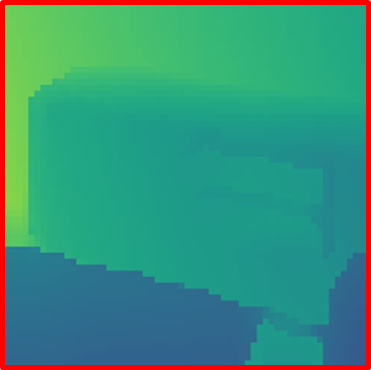}}
	\hfill
	\subfigure{\includegraphics[width=\imgwidth]{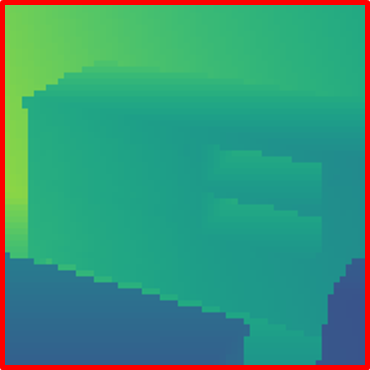}}
	
	\vspace{\largespace}	
	
	\renewcommand{\fignum}{5865}
	\subfigure{\includegraphics[width=\imgwidth]{\dirname\fignum/img_0.png}}
	\hfill
	\subfigure{\includegraphics[width=\imgwidth]{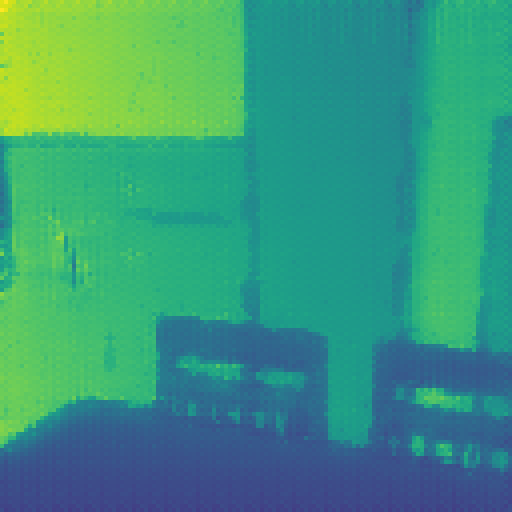}}
	\hfill
	\subfigure{\includegraphics[width=\imgwidth]{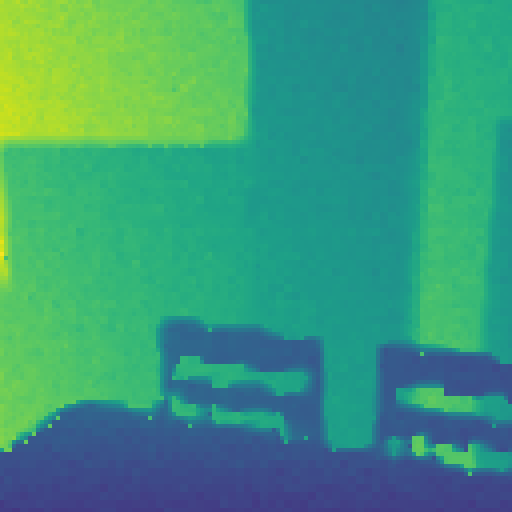}}
	\hfill
	\subfigure{\includegraphics[width=\imgwidth]{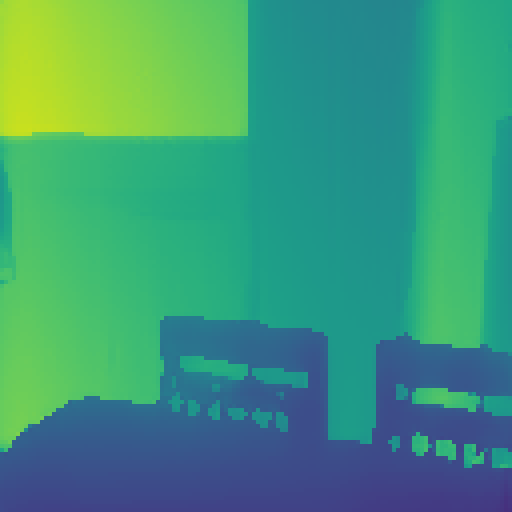}}
	\hfill
	\subfigure{\includegraphics[width=\imgwidth]{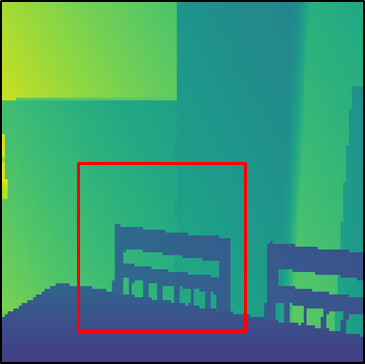}}
	
	\vspace{\smallspace}
	
	\subfigure{\stackunder[4pt]{\includegraphics[width=\imgwidth]{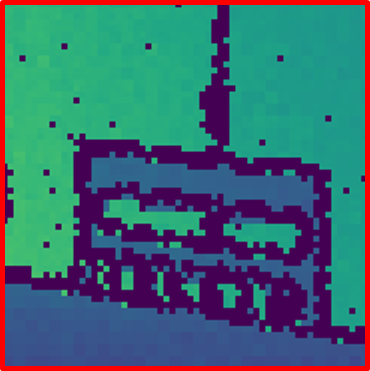}}{\small Input}}
	\hfill
	\subfigure{\stackunder[4pt]{\includegraphics[width=\imgwidth]{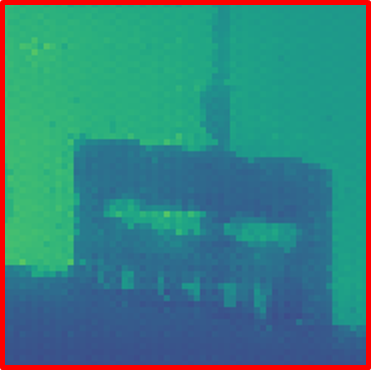}}{\small Cyc. GAN}}
	\hfill
	\subfigure{\stackunder[4pt]{\includegraphics[width=\imgwidth]{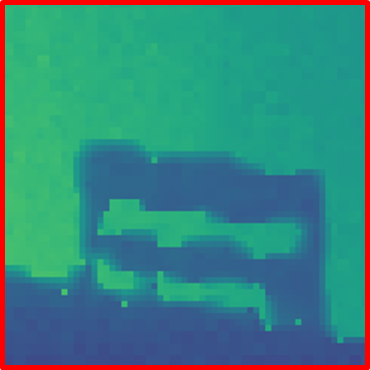}}{\small SDS~\cite{sparse-depth-sensing}}}
	\hfill
	\subfigure{\stackunder[4pt]{\includegraphics[width=\imgwidth]{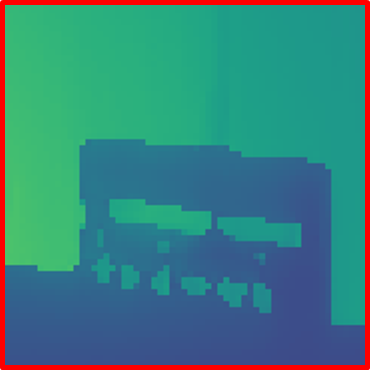}}{\small \textbf{Ours}}}
	\hfill
	\subfigure{\stackunder[4pt]{\includegraphics[width=\imgwidth]{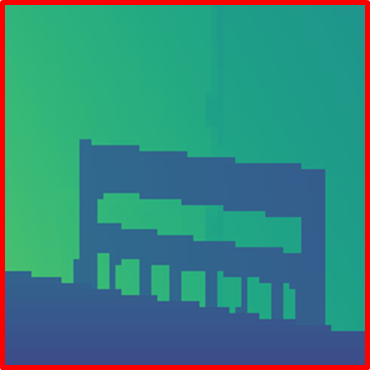}}{\small GT}}
	
	\caption{Example enhancement results on the synthetic dataset.}
	\label{fig:synthetic-comparison}
\end{figure}

Table~(\ref{tbl:synthetic-results-16}) lists the quantitative results of the synthetic experiment. We use $b=16$ and $s=4$ in the PNCC computation, though we note that the results behave similarly across different block sizes and steps. 
Example results are shown in Figure~(\ref{fig:synthetic-comparison}).
As can be seen, the quantitative results indicate that our method is indeed recovering more detail than the alternatives. Examining the images, we see that the modified GAN formulation produces much sharper and more detailed images than either the original Cycle-GAN or the sparse sensing algorithm, in-line with the local correlation metric.

\begin{table}
	\vspace{0.07in} 
	\renewcommand\arraystretch{1}
	\setlength{\tabcolsep}{6pt}
	\centering
	\begin{tabular}{|c|c|c|c|c|}
		\hline
		\textbf{Base} & \textbf{Improved Net} & \textbf{Tri-Cycle} & \textbf{Sparse Sensing} \\
		\hline
		0.668 & 0.869 & \textbf{0.879} & 0.736 \\
		\hline
	\end{tabular}
	\caption{Quantitative results of the synthetic experiment in terms of PNCC. Left to right: base Cycle-GAN, Cycle-GAN with improved generator, our result with Tri-Cycle loss, result of~\cite{sparse-depth-sensing}.}
	\label{tbl:synthetic-results-16}
\end{table}

\subsection{Experiments with Real Depth Data}
\label{sec:office-res}

To demonstrate our method in more real conditions, we created a dataset of real-world depth images captured in an office setting. The images were captured independently using the RealSense and Kinect~2 cameras, with no synchronization between them. After some basic filtering (e.g., removing similar images or images with very little content) we arrived at a dataset consisting of just over 1,000 images from each camera. We note that due to the unconstrained manner in which this dataset was captured, we have no ground-truth for these images, and thus can only perform a qualitative evaluation of the results. On the other hand, the construction of this dataset makes it truly unsupervised, and thus well-representative of a real-world scenario. Figure~(\ref{fig:kinect-rs}) shows an example from this dataset, with additional examples provided in the results figures.

\begin{figure}[t]
	\centering
	\renewcommand{\dirname}{./figures/ablationA/BtoA_example_}
	\renewcommand{\largespace}{-0.08in}
	\renewcommand{\imgwidth}{0.19\linewidth}
	
	\renewcommand{\fignum}{5}
	\subfigure{\includegraphics[width=\imgwidth]{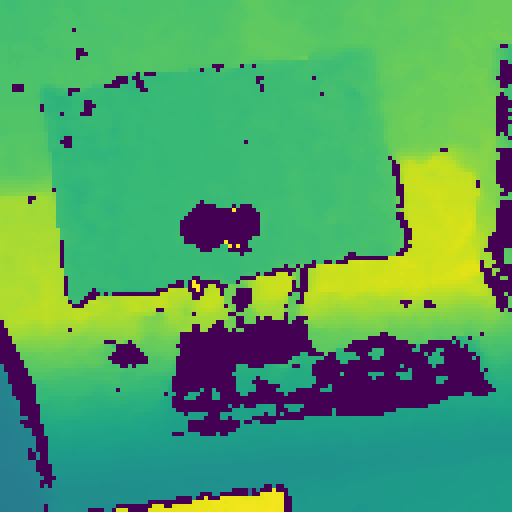}}
	\hfill
	\subfigure{\includegraphics[width=\imgwidth]{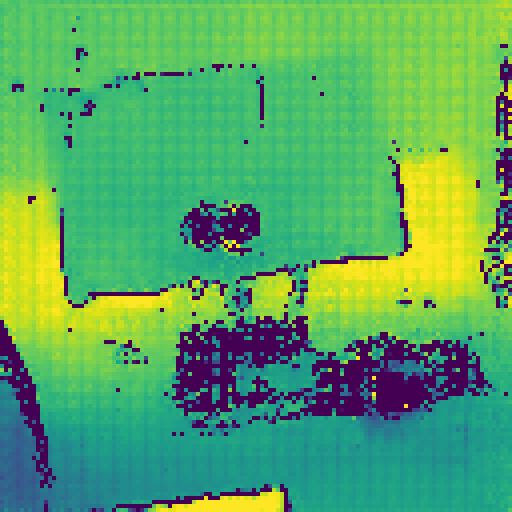}}
	\hfill
	\subfigure{\includegraphics[width=\imgwidth]{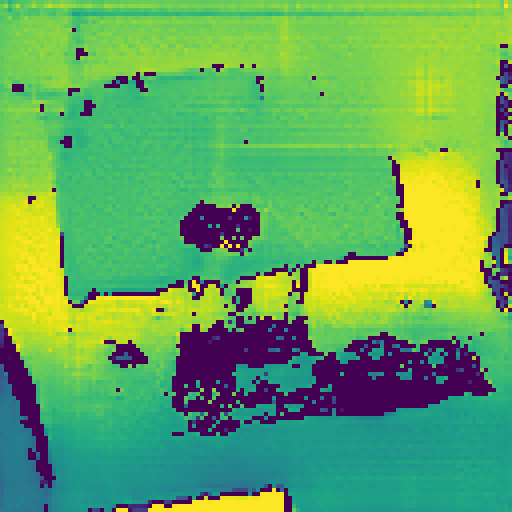}}
	\hfill
	\subfigure{\includegraphics[width=\imgwidth]{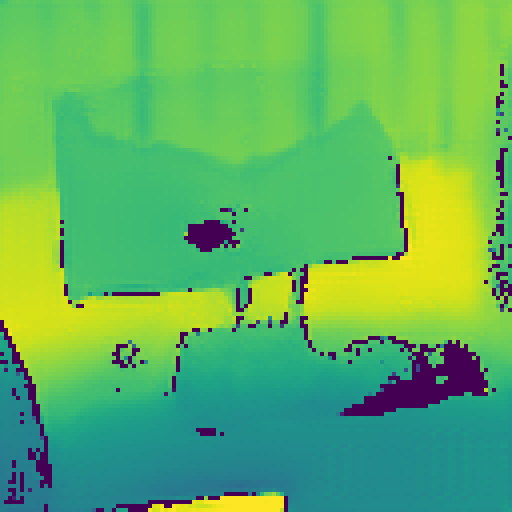}}
	\hfill
	\subfigure{\includegraphics[width=\imgwidth]{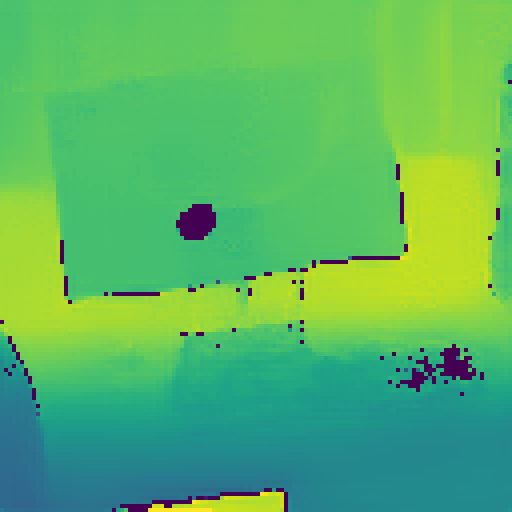}}
	
	\vspace{\largespace}
	
	\renewcommand{\fignum}{821}
	\subfigure{\includegraphics[width=\imgwidth]{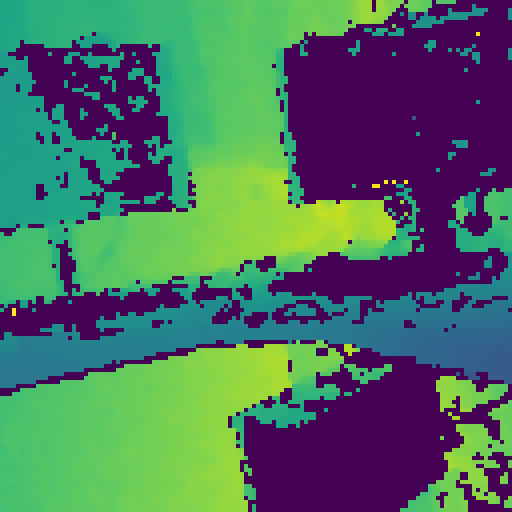}}
	\hfill
	\subfigure{\includegraphics[width=\imgwidth]{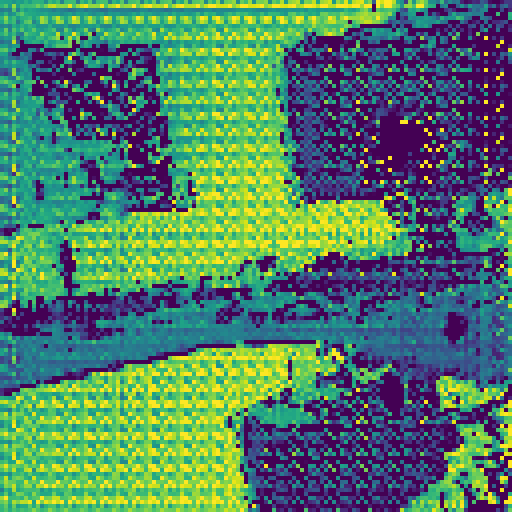}}
	\hfill
	\subfigure{\includegraphics[width=\imgwidth]{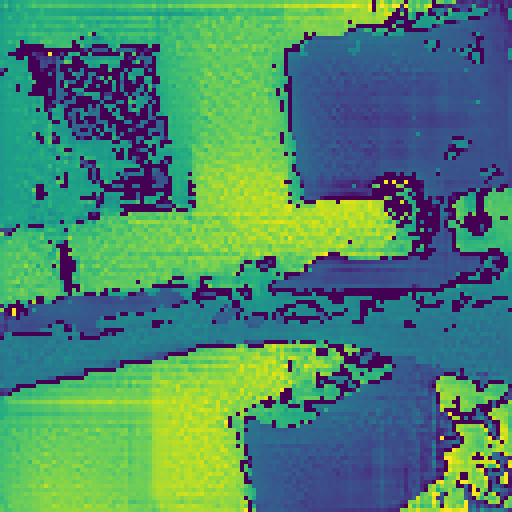}}
	\hfill
	\subfigure{\includegraphics[width=\imgwidth]{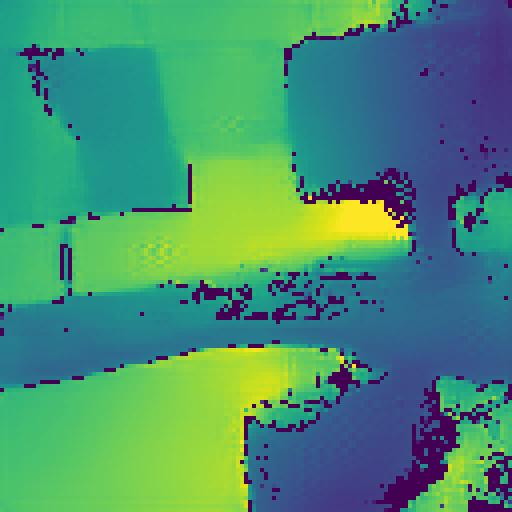}}
	\hfill
	\subfigure{\includegraphics[width=\imgwidth]{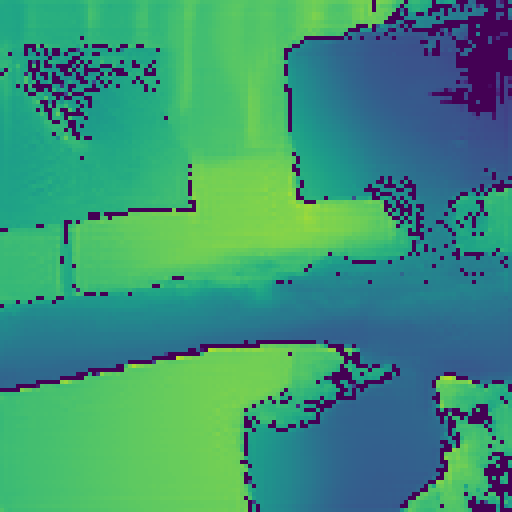}}
	
	\vspace{\largespace}
	
	\renewcommand{\fignum}{772}
	\subfigure{\includegraphics[width=\imgwidth]{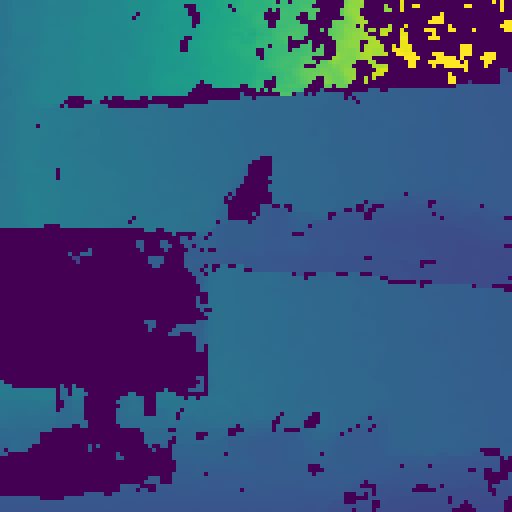}}
	\hfill
	\subfigure{\includegraphics[width=\imgwidth]{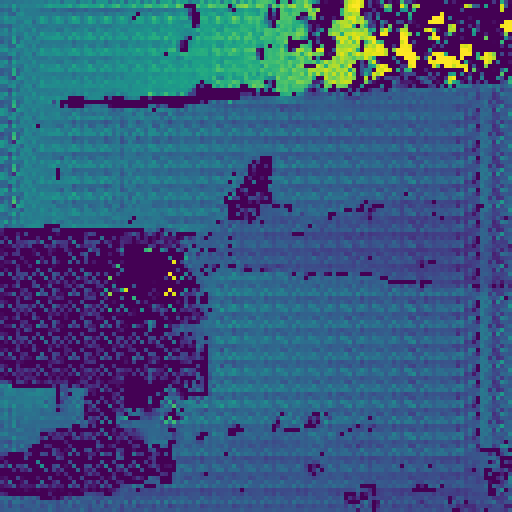}}
	\hfill
	\subfigure{\includegraphics[width=\imgwidth]{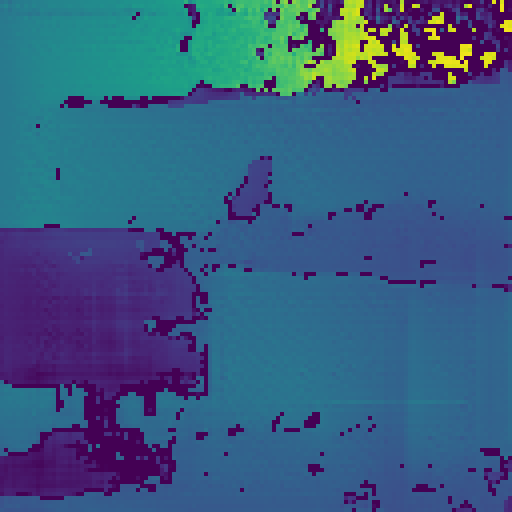}}
	\hfill
	\subfigure{\includegraphics[width=\imgwidth]{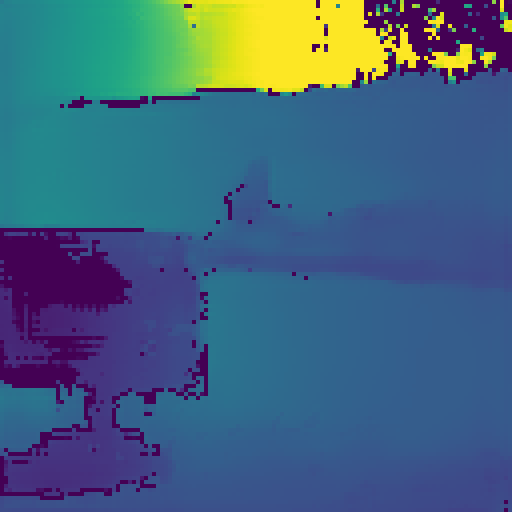}}
	\hfill
	\subfigure{\includegraphics[width=\imgwidth]{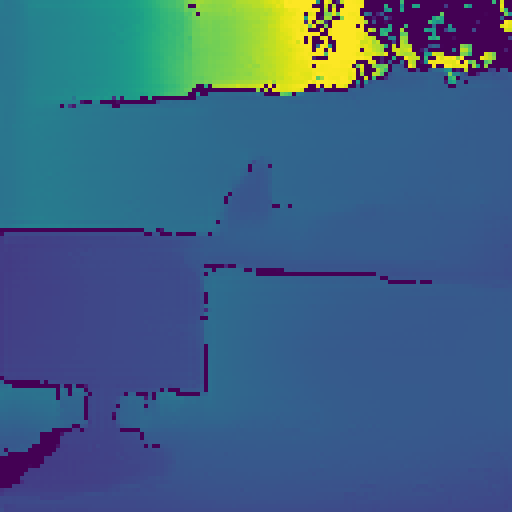}}
	
	\vspace{\largespace}
	
	\renewcommand{\fignum}{412}
	\subfigure{\includegraphics[width=\imgwidth]{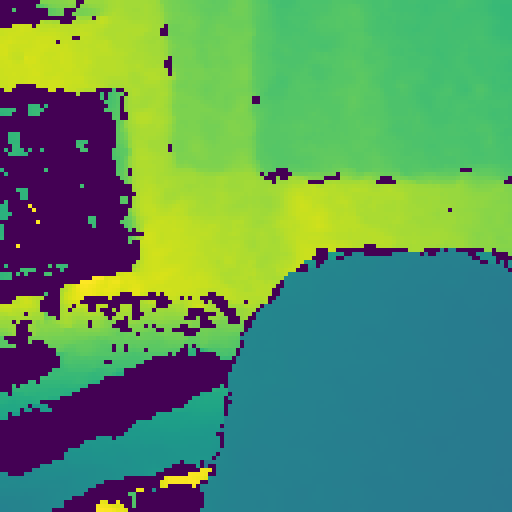}}
	\hfill
	\subfigure{\includegraphics[width=\imgwidth]{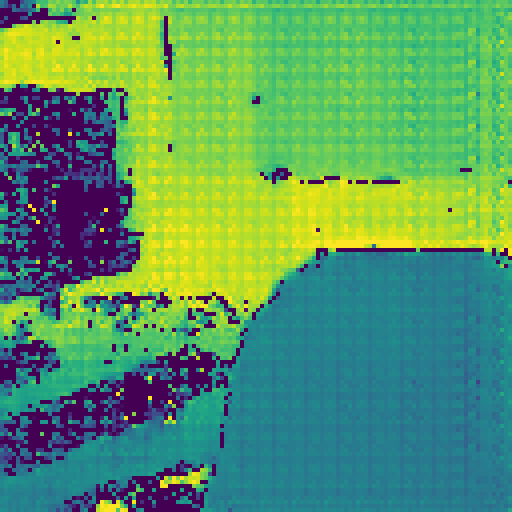}}
	\hfill
	\subfigure{\includegraphics[width=\imgwidth]{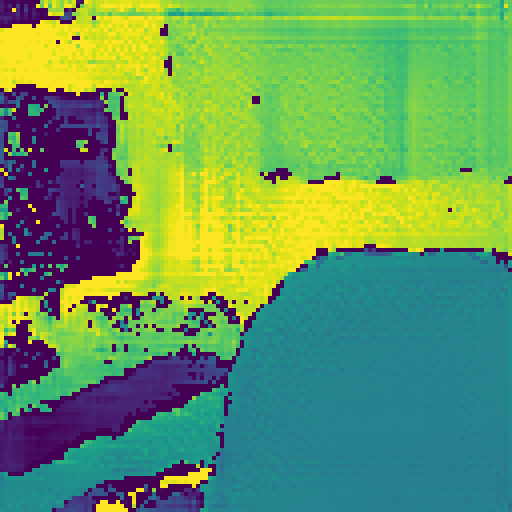}}
	\hfill
	\subfigure{\includegraphics[width=\imgwidth]{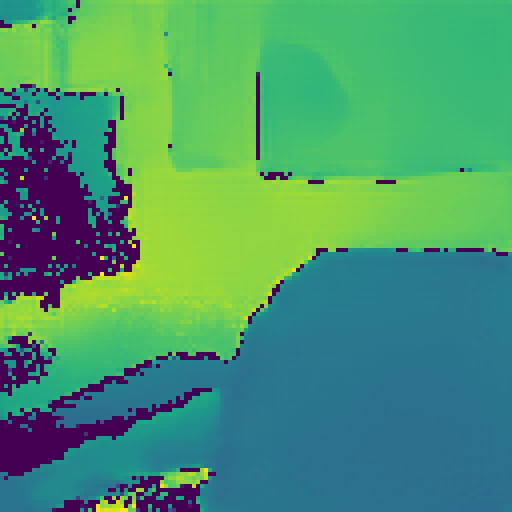}}
	\hfill
	\subfigure{\includegraphics[width=\imgwidth]{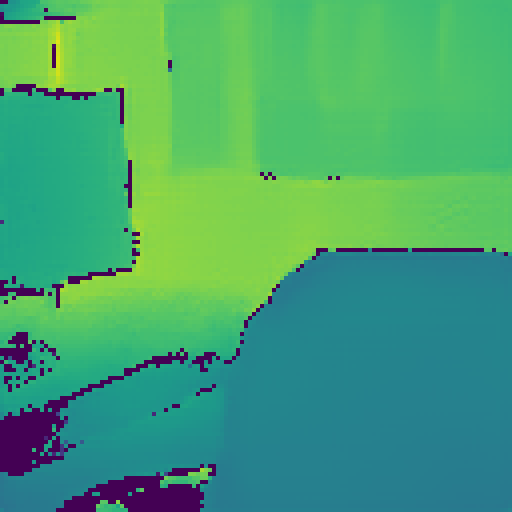}}
	
	\caption{Effect of generator architecture. Left to right: Input image; result with original architecture; result with increased number of channels; result with increased number of layers; result with both modifications.}
	\label{fig:ablation-a}
\end{figure}

Figure~(\ref{fig:ablation-a}) demonstrates the effects of the generator architecture on the results. To isolate the parameters for this experiment, we do not employ any of the new losses in this case, and only vary the network architecture.  We consider the following architectures: (1) original network; (2) original network with increased number of channels; (3) original network with increased number of layers; and (4) the final network. As can be seen, the original architecture is essentially unusable for this task, producing strong artifacts and providing no visible enhancement. Increasing the number of channels or the number of layers each have a notable effect in terms of reducing artifacts and filling-in holes, though with limited success. 
Combining both modifications produces the best results, with the fewest visible artifacts and the most accurate hole filling. It is thus clear that both width and depth are crucial for handling the challenges of low-quality depth. We note that in particular, the increased number of channels in the earlier network layers deviates from the standard practice for CNNs~\cite{vgg}, though proves advantageous in this case.

\begin{figure}
	\centering
	\renewcommand{\dirname}{./figures/tricycle6/BtoA_example_}
	\renewcommand{\imgwidth}{0.19\linewidth}
	\renewcommand{\largespace}{-0.08in}
	
	\renewcommand{\fignum}{155}
	\subfigure{\includegraphics[width=\imgwidth]{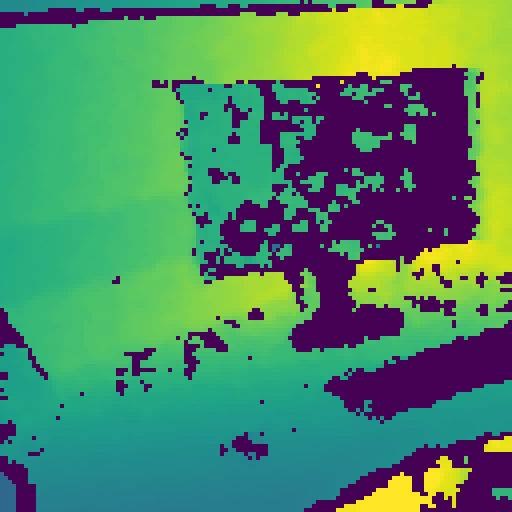}}
	\hfill
	\subfigure{\includegraphics[width=\imgwidth]{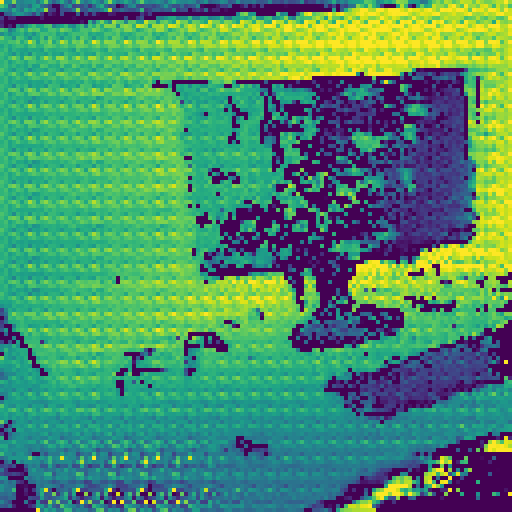}}
	\hfill
	\subfigure{\includegraphics[width=\imgwidth]{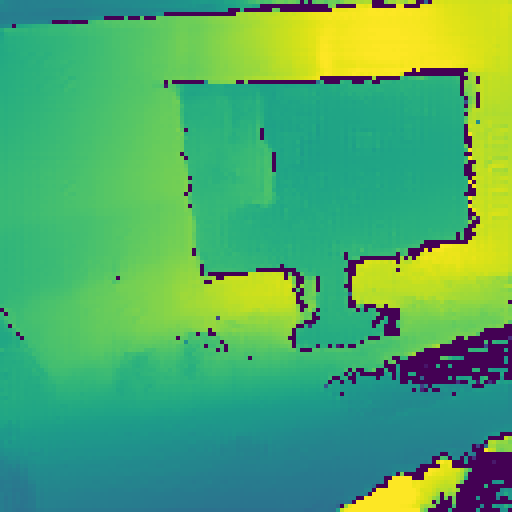}}
	\hfill
	\subfigure{\includegraphics[width=\imgwidth]{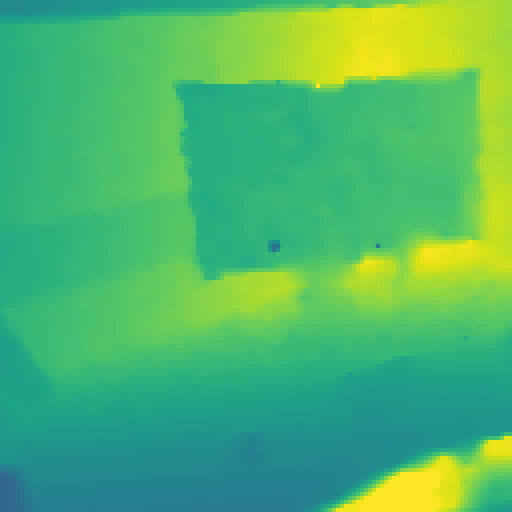}}
	\hfill
	\subfigure{\includegraphics[width=\imgwidth]{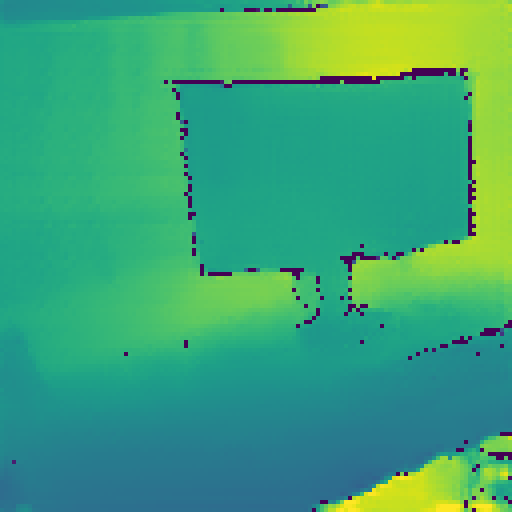}}
	
	\vspace{\largespace}
	
	\renewcommand{\fignum}{425}
	\subfigure{\includegraphics[width=\imgwidth]{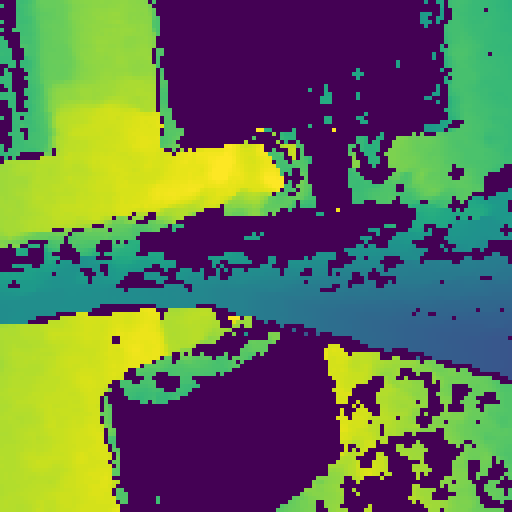}}
	\hfill
	\subfigure{\includegraphics[width=\imgwidth]{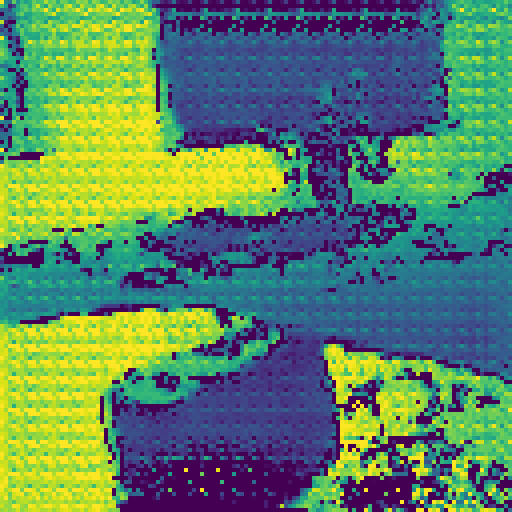}}
	\hfill
	\subfigure{\includegraphics[width=\imgwidth]{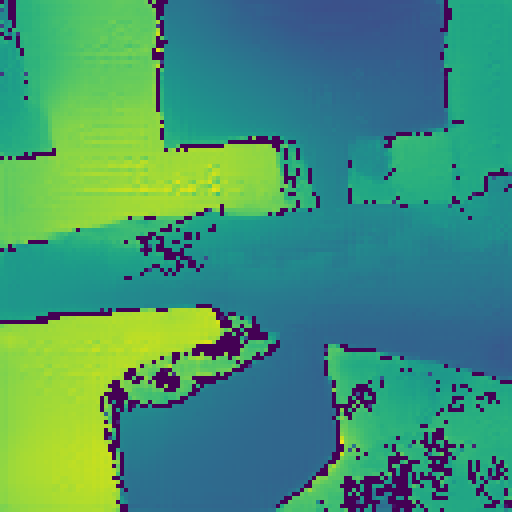}}
	\hfill
	\subfigure{\includegraphics[width=\imgwidth]{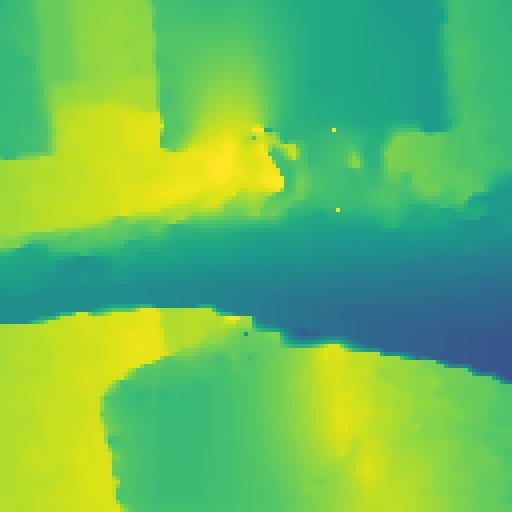}}
	\hfill
	\subfigure{\includegraphics[width=\imgwidth]{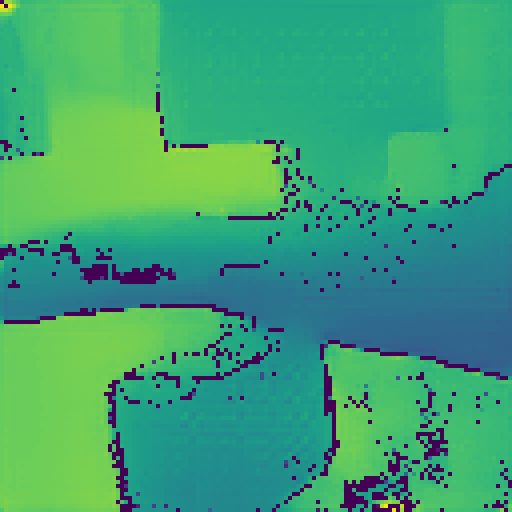}}
	
	\vspace{\largespace}
	
	\renewcommand{\fignum}{412}
	\subfigure{\includegraphics[width=\imgwidth]{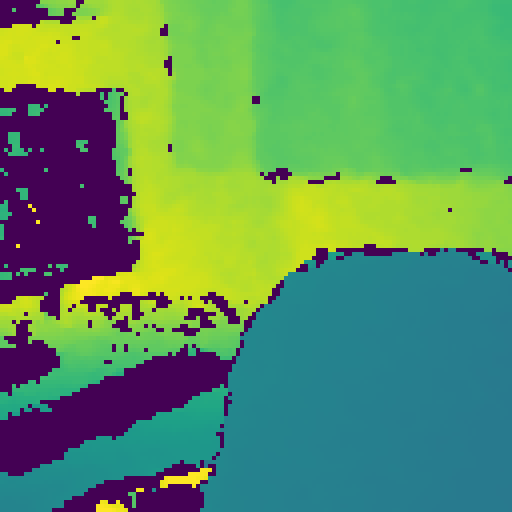}}
	\hfill
	\subfigure{\includegraphics[width=\imgwidth]{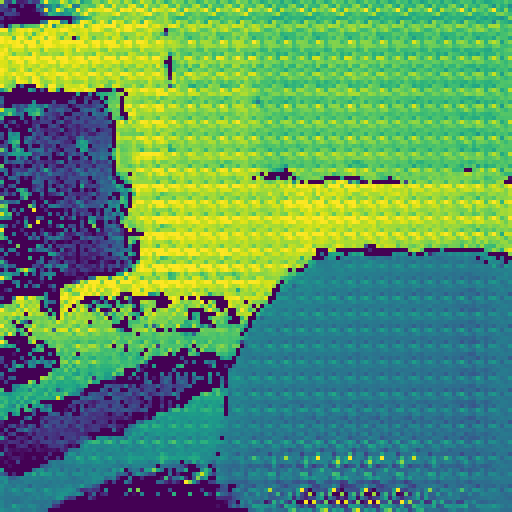}}
	\hfill
	\subfigure{\includegraphics[width=\imgwidth]{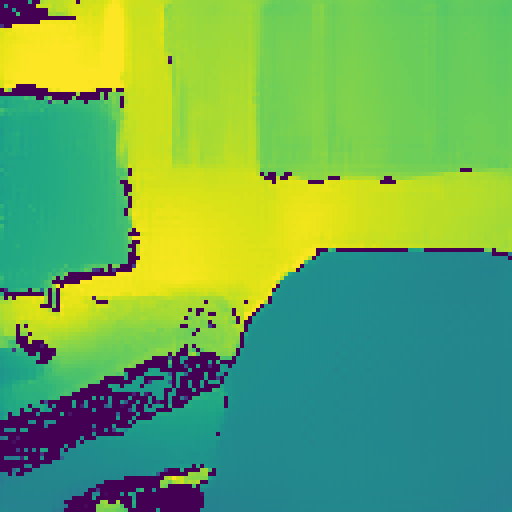}}
	\hfill
	\subfigure{\includegraphics[width=\imgwidth]{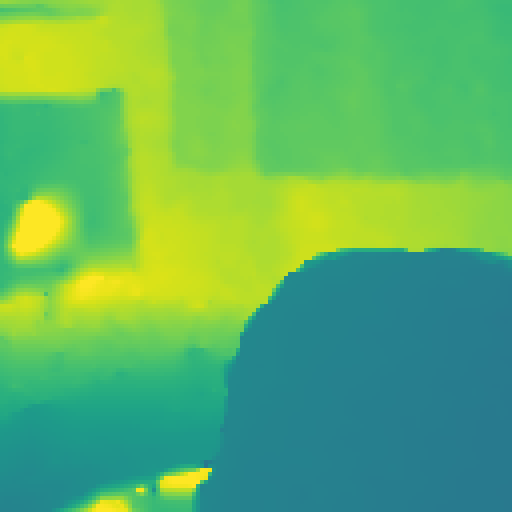}}
	\hfill
	\subfigure{\includegraphics[width=\imgwidth]{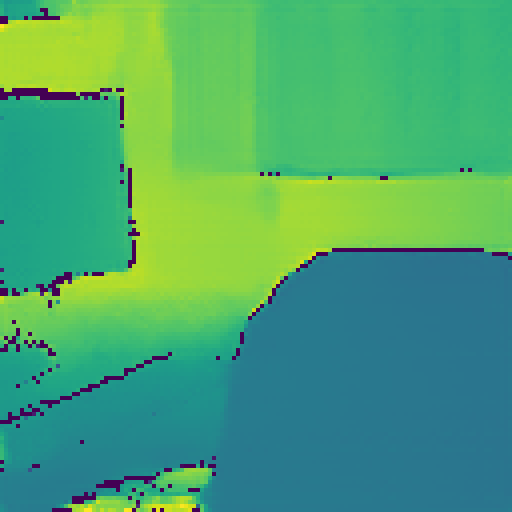}}
	
	\vspace{\largespace}
	
	\renewcommand{\fignum}{794}
	\subfigure{\includegraphics[width=\imgwidth]{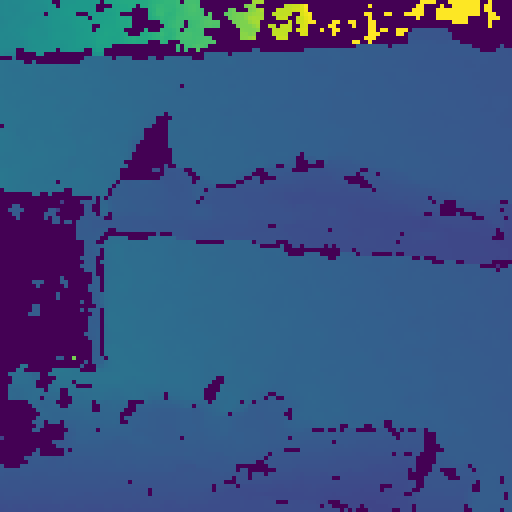}}
	\hfill
	\subfigure{\includegraphics[width=\imgwidth]{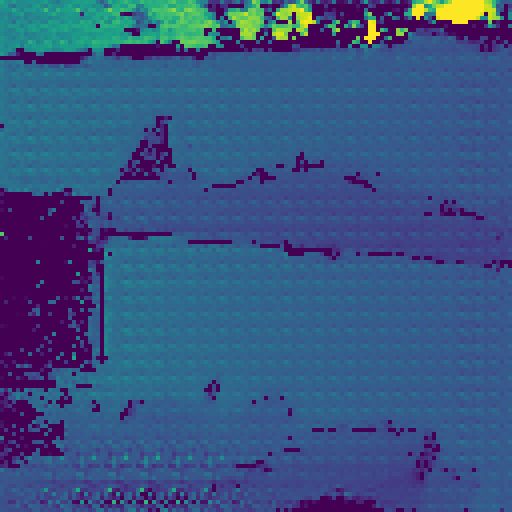}}
	\hfill
	\subfigure{\includegraphics[width=\imgwidth]{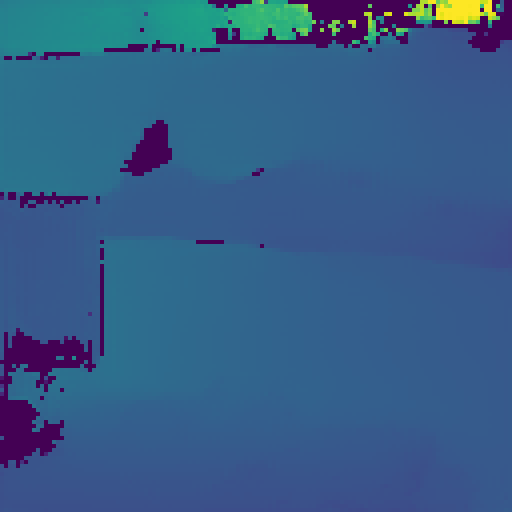}}
	\hfill
	\subfigure{\includegraphics[width=\imgwidth]{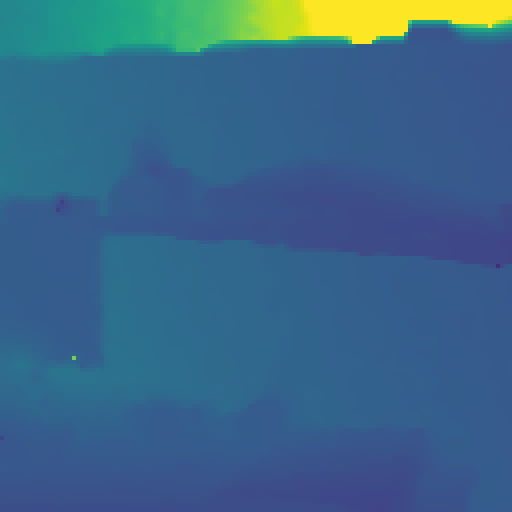}}
	\hfill
	\subfigure{\includegraphics[width=\imgwidth]{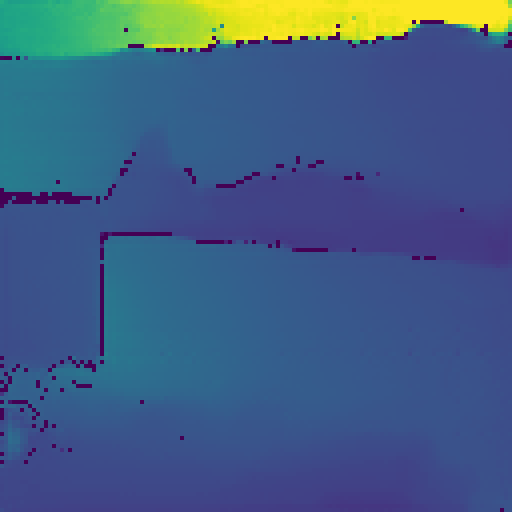}}
	
	\vspace{\largespace}
		
	\renewcommand{\fignum}{450}
	\subfigure{\includegraphics[width=\imgwidth]{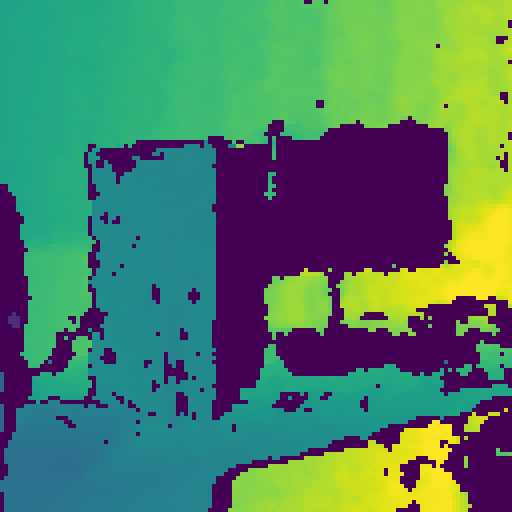}}
	\hfill
	\subfigure{\includegraphics[width=\imgwidth]{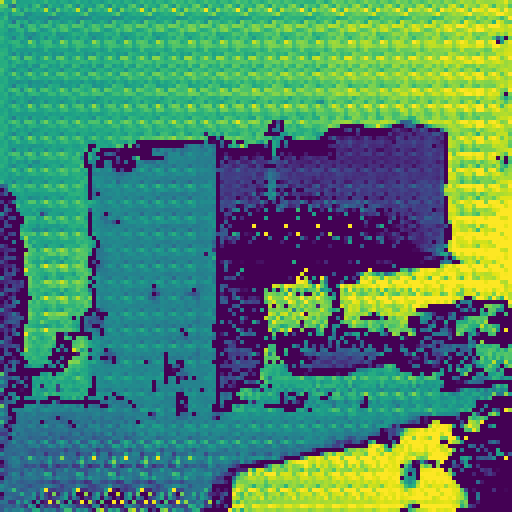}}
	\hfill
	\subfigure{\includegraphics[width=\imgwidth]{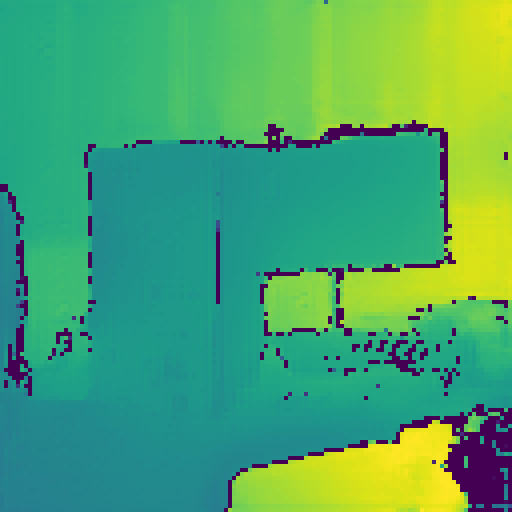}}
	\hfill
	\subfigure{\includegraphics[width=\imgwidth]{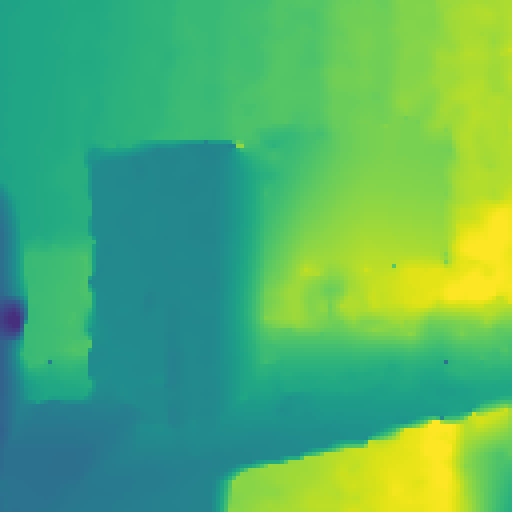}}
	\hfill
	\subfigure{\includegraphics[width=\imgwidth]{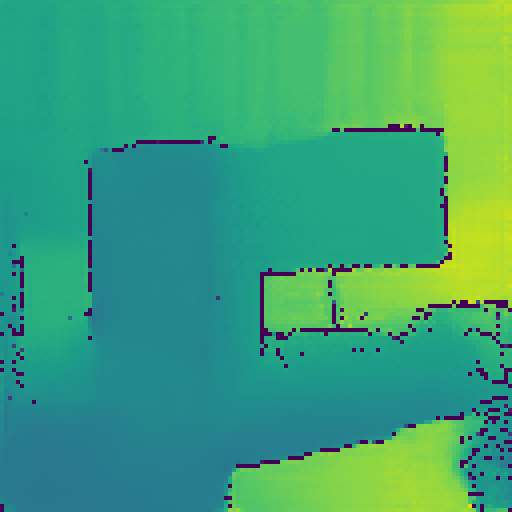}}
	
	\vspace{\largespace}

	\renewcommand{\fignum}{32}
	\subfigure{\includegraphics[width=\imgwidth]{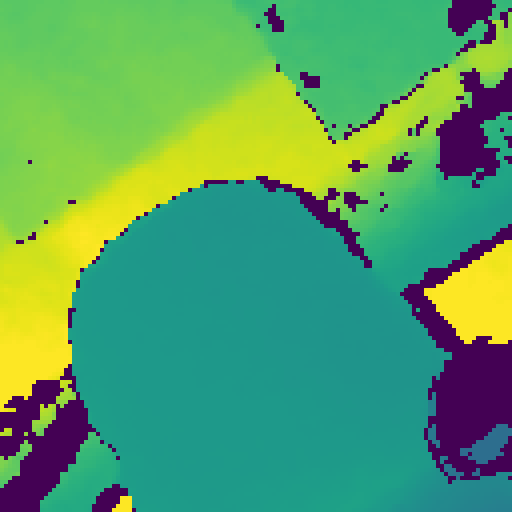}}
	\hfill
	\subfigure{\includegraphics[width=\imgwidth]{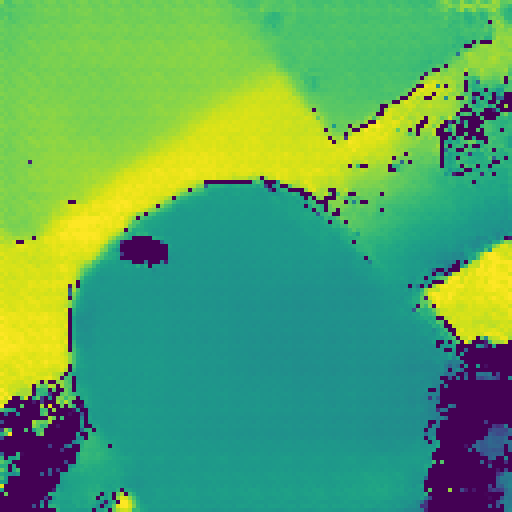}}
	\hfill
	\subfigure{\includegraphics[width=\imgwidth]{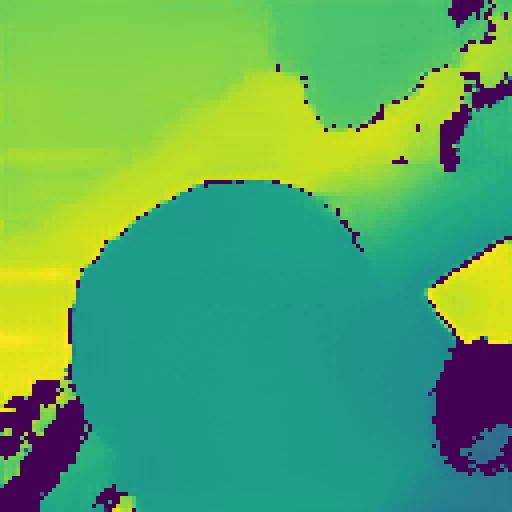}}
	\hfill
	\subfigure{\includegraphics[width=\imgwidth]{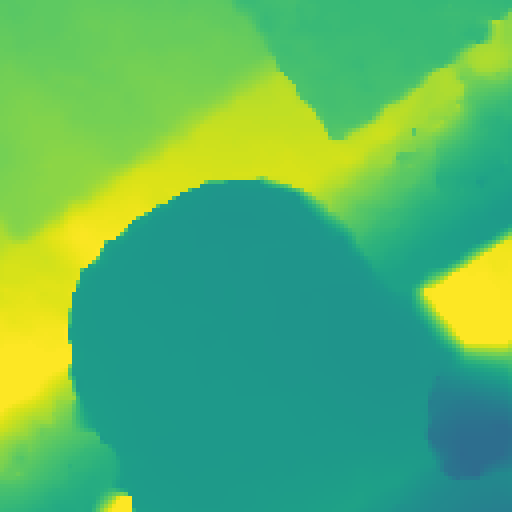}}
	\hfill
	\subfigure{\includegraphics[width=\imgwidth]{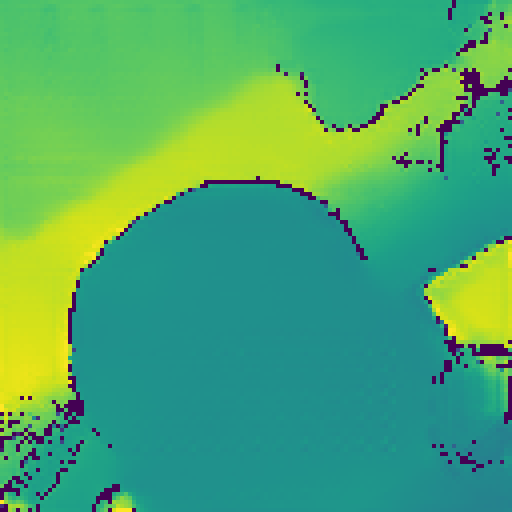}}
	
	\vspace{\largespace}
	
	\renewcommand{\fignum}{751}
	\subfigure{\stackunder[4pt]{\includegraphics[width=\imgwidth]{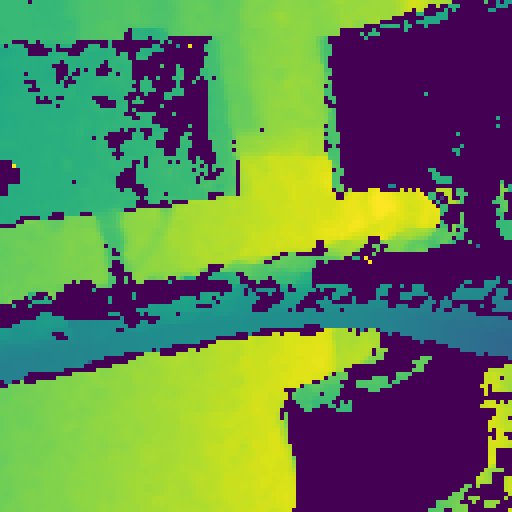}}{\small Input}}
	\hfill
	\subfigure{\stackunder[4pt]{\includegraphics[width=\imgwidth]{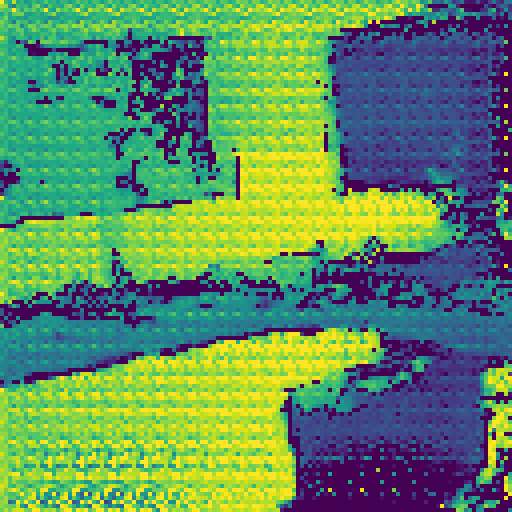}}{\small Cyc. GAN}}
	\hfill
	\subfigure{\stackunder[4pt]{\includegraphics[width=\imgwidth]{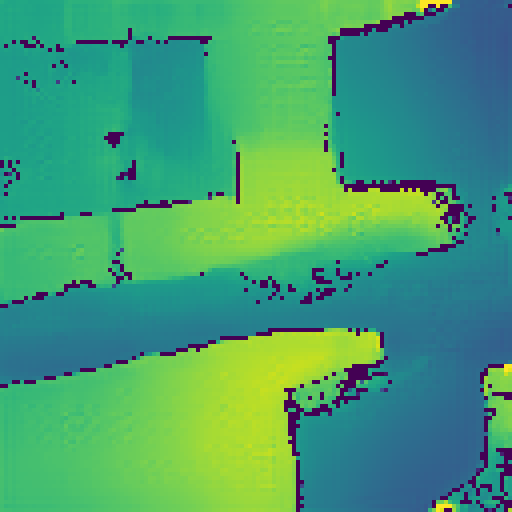}}{\small New Net}}
	\hfill
	\subfigure{\stackunder[4pt]{\includegraphics[width=\imgwidth]{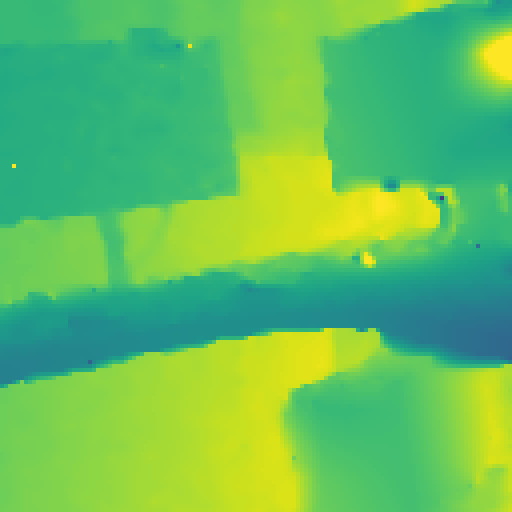}}{\small SDS~\cite{sparse-depth-sensing}}}
	\hfill
	\subfigure{\stackunder[4pt]{\includegraphics[width=\imgwidth]{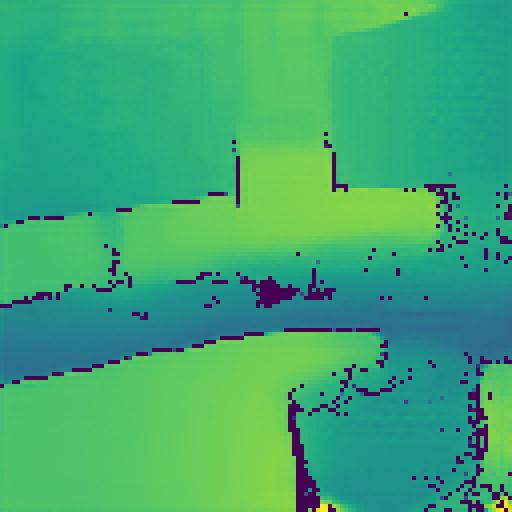}}{\small \textbf{Ours}}}
	
	\caption{Example results for the \emph{Office} dataset.}
	\label{fig:results-tricycle-1}
\end{figure}

Finally, Figure~(\ref{fig:results-tricycle-1}) shows some recovery results of our full Tri-Cycle GAN framework.
As before, we compare our results to those of Sparse Depth Sensing~\cite{sparse-depth-sensing}. We also show results with and without the Tri-Cycle loss, to demonstrate its effects on recovery performance. As can be seen, the method~\cite{sparse-depth-sensing} struggles with these images, exhibiting over-smoothness, jagged object edges, and intensification of outlier pixels leading to unnatural holes in objects. Clearly, the degradation model assumed by this method is too simplistic for this task. Continuing with the Cycle-GAN, increasing the network size has a significant effect on the results, though the output still suffers from visible artifacts and missing regions. Adding the Tri-Cycle loss leads to a notable improvement in the results, producing more realistic and detailed images with fewer artifacts and missing pixels. Indeed, as many of these artifacts are in regions which were strongly corrupted in the input image, we attribute these improvements to the Tri-Cycle loss, which relaxes the requirement to recover the exact degraded input by the inverse translation.

\subsection{Experiments with the DROT Dataset}
\label{sec:drot-res}

\begin{figure}
	\centering
	\renewcommand{\dirname}{./figures/drot-dataset/img1}
	\renewcommand{\imgwidth}{0.28\linewidth}
	\subfigure{\stackunder[4pt]{\includegraphics[width=\imgwidth]{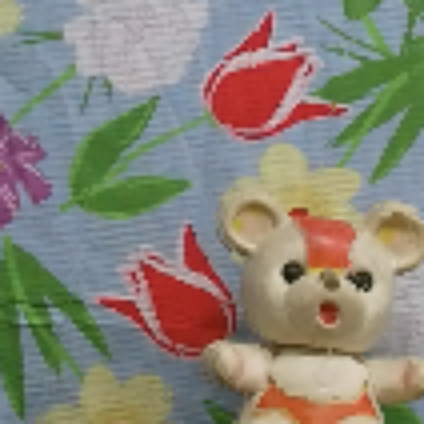}}{\small Color}}
	\hfill
	\subfigure{\stackunder[4pt]{\includegraphics[width=\imgwidth]{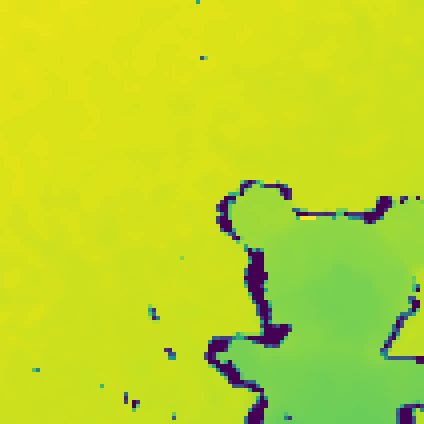}}{\small RealSense}}
	\hfill
	\subfigure{\stackunder[4pt]{\includegraphics[width=\imgwidth]{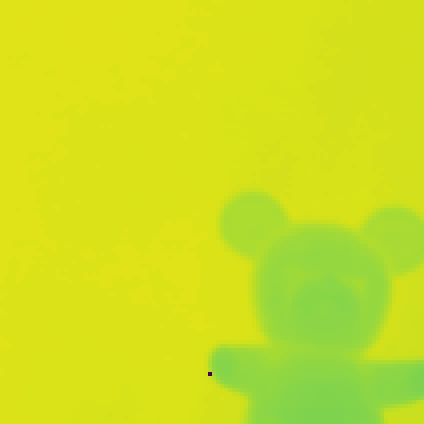}}{\small Kinect 2}}
	
	\caption{Example aligned image set from the DROT dataset.}
	\label{fig:daniel-image-set}
\end{figure}

The Depth Restoration Occlusionless Temporal dataset, or \emph{DROT}~\cite{daniel-kinect-rs}, is a carefully captured and post-processed set of RealSense and Kinect~2 images, which are nearly pixel-level aligned.~\footnote{The dataset also includes color, Kinect~1, and 3D~DAVID images, though we do not use these in this work.} The dataset consists of~112 image sets which have been recorded in a studio setting, employing a highly accurate calibration process between the cameras. Figure~(\ref{fig:daniel-image-set}) shows an example from this dataset.
We use this dataset to quantify the performance of our method on actual RealSense depth maps. It should be noted, though, that due to the controlled environment and specifically-chosen scene and materials, this dataset exhibits much lighter degradations than those our method was intended to handle.

\begin{figure}
	\centering
	\renewcommand{\dirname}{./figures/daniel-results/BtoA_example_}
	\renewcommand{\imgwidth}{0.19\linewidth}
	\renewcommand{\largespace}{-0.08in}
	
	\renewcommand{\fignum}{4}
	\subfigure{\includegraphics[width=\imgwidth]{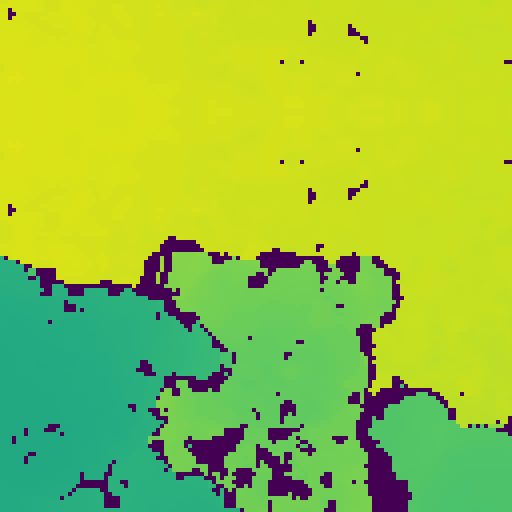}}
	\hfill
	\subfigure{\includegraphics[width=\imgwidth]{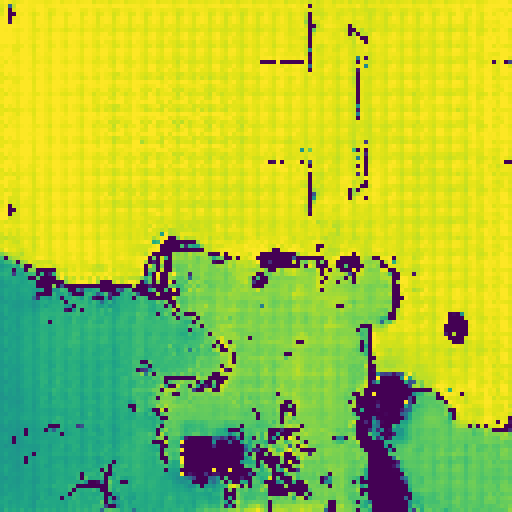}}
	\hfill
	\subfigure{\includegraphics[width=\imgwidth]{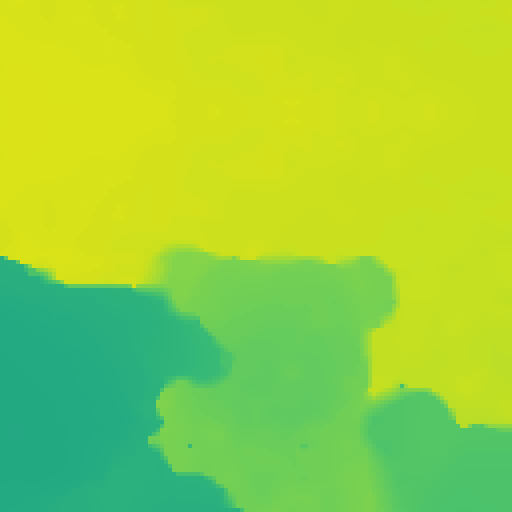}}
	\hfill
	\subfigure{\includegraphics[width=\imgwidth]{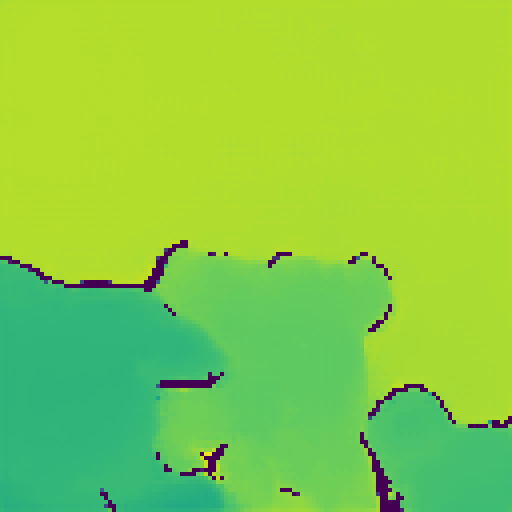}}
	\hfill
	\subfigure{\includegraphics[width=\imgwidth]{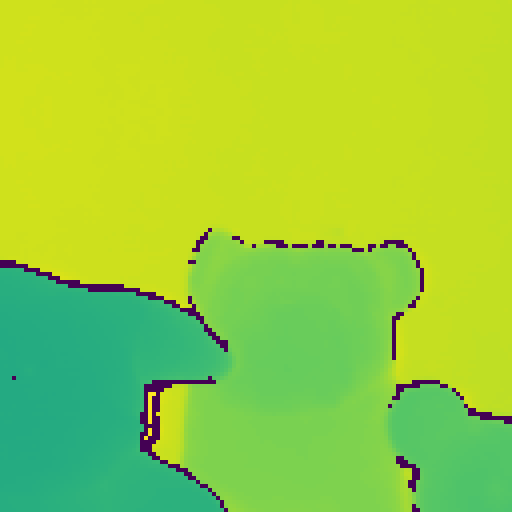}}
	
	\vspace{\largespace}
	
	\renewcommand{\fignum}{41}
	\subfigure{\includegraphics[width=\imgwidth]{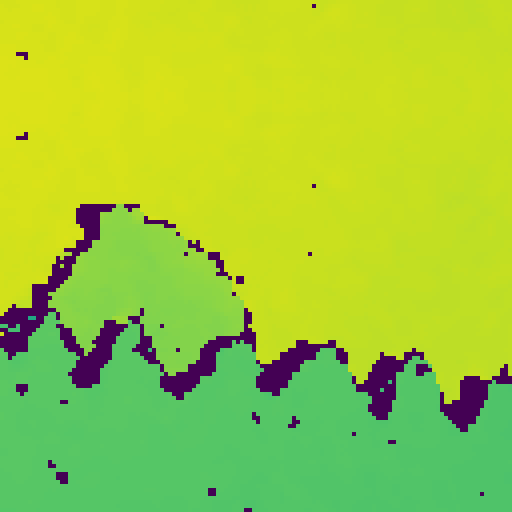}}
	\hfill
	\subfigure{\includegraphics[width=\imgwidth]{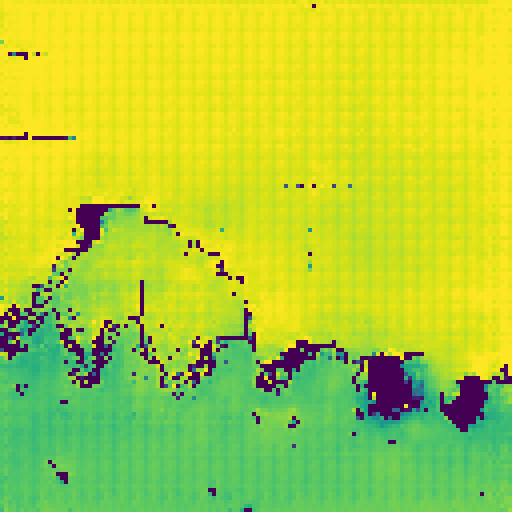}}
	\hfill
	\subfigure{\includegraphics[width=\imgwidth]{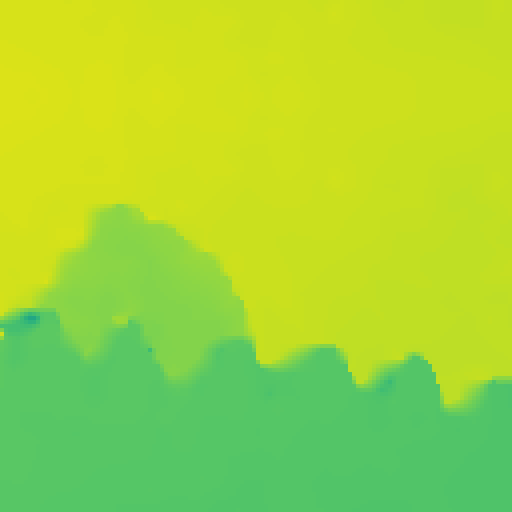}}
	\hfill
	\subfigure{\includegraphics[width=\imgwidth]{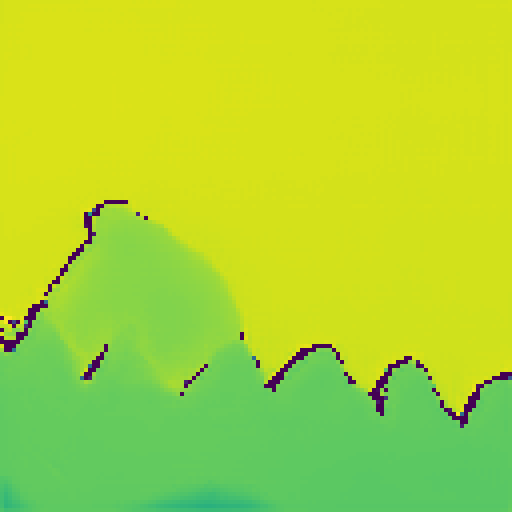}}
	\hfill
	\subfigure{\includegraphics[width=\imgwidth]{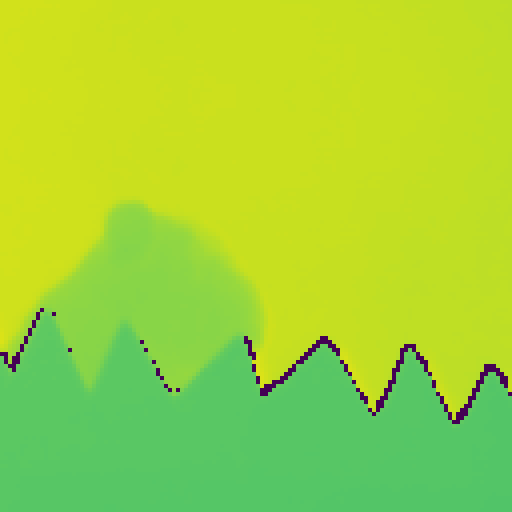}}
	
	\vspace{\largespace}
	
	\renewcommand{\fignum}{3}
	\subfigure{\includegraphics[width=\imgwidth]{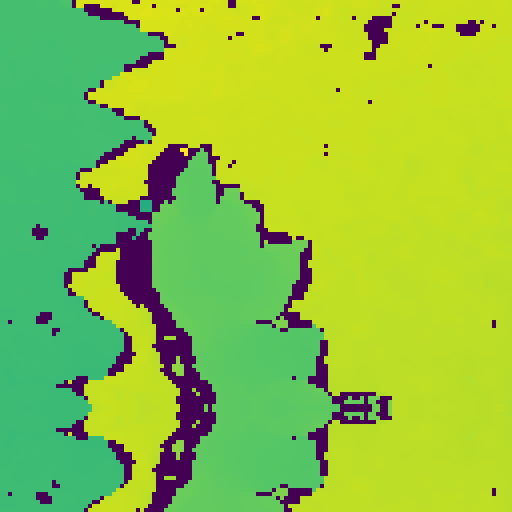}}
	\hfill
	\subfigure{\includegraphics[width=\imgwidth]{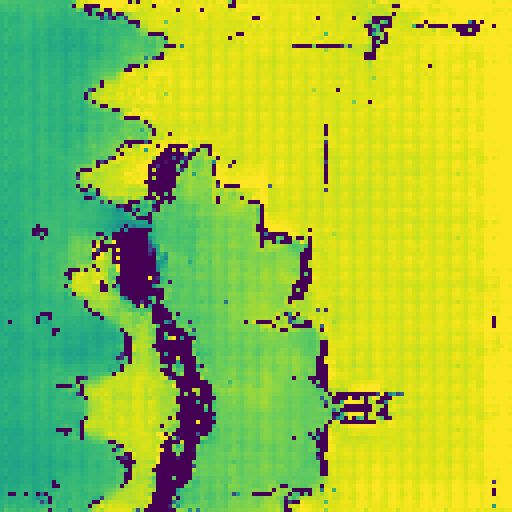}}
	\hfill
	\subfigure{\includegraphics[width=\imgwidth]{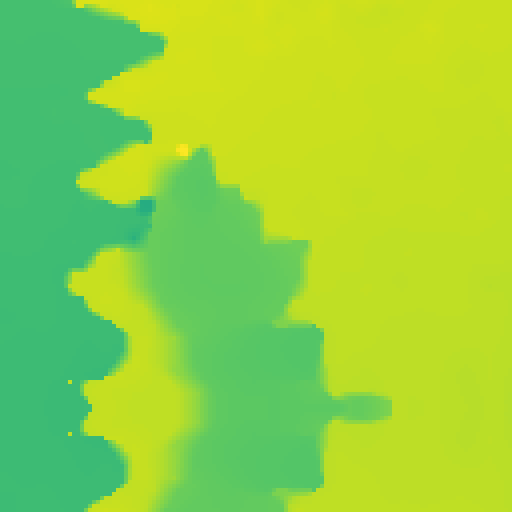}}
	\hfill
	\subfigure{\includegraphics[width=\imgwidth]{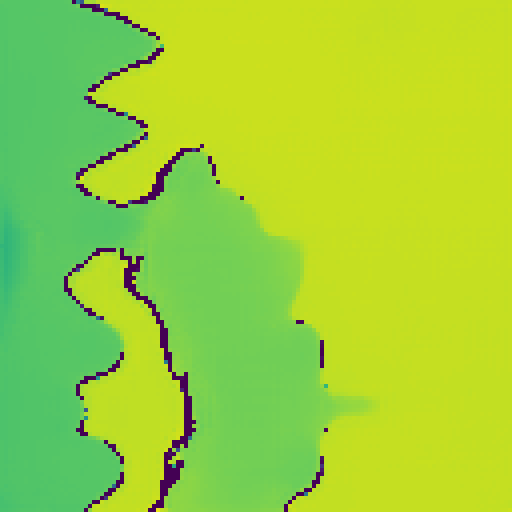}}
	\hfill
	\subfigure{\includegraphics[width=\imgwidth]{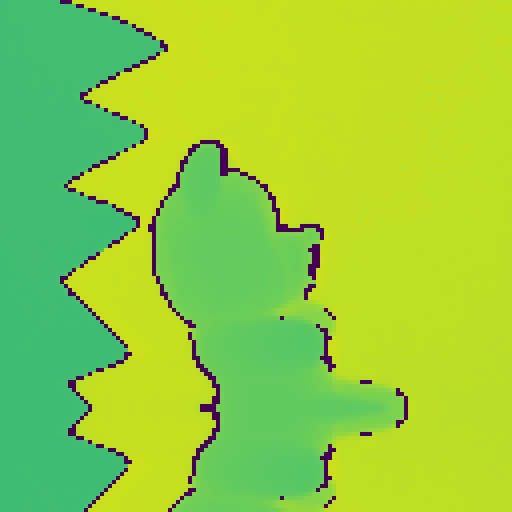}}
	
	\vspace{\largespace}
	
	\renewcommand{\fignum}{71}
	\subfigure{\includegraphics[width=\imgwidth]{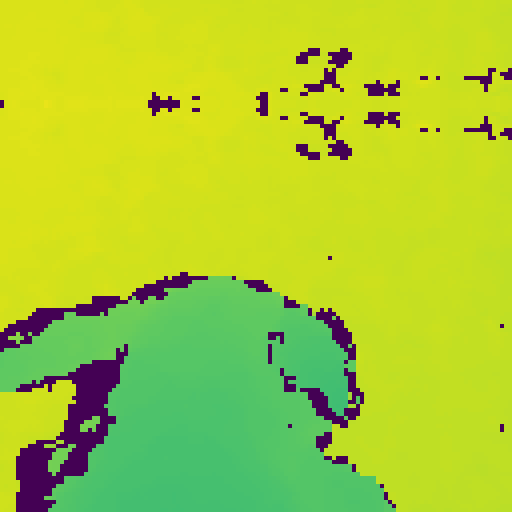}}
	\hfill
	\subfigure{\includegraphics[width=\imgwidth]{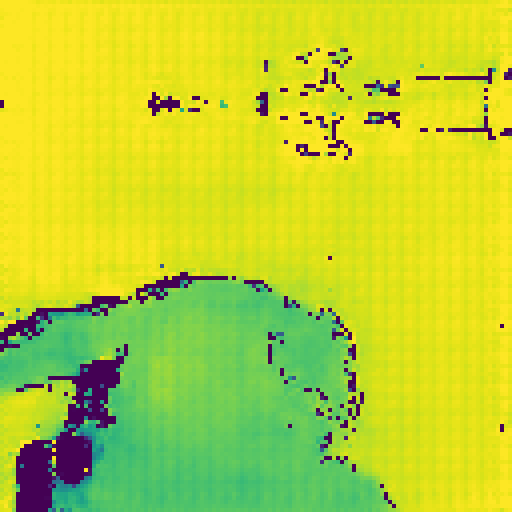}}
	\hfill
	\subfigure{\includegraphics[width=\imgwidth]{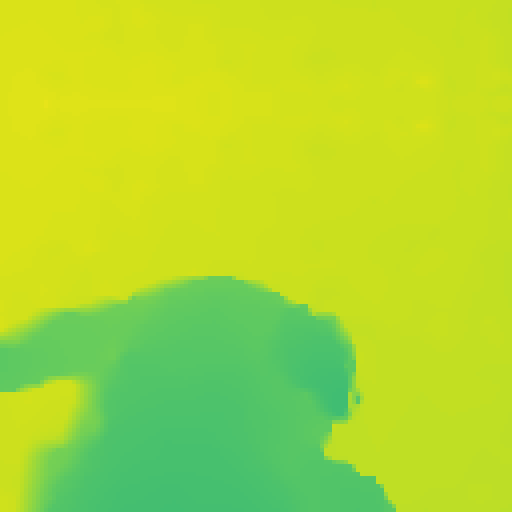}}
	\hfill
	\subfigure{\includegraphics[width=\imgwidth]{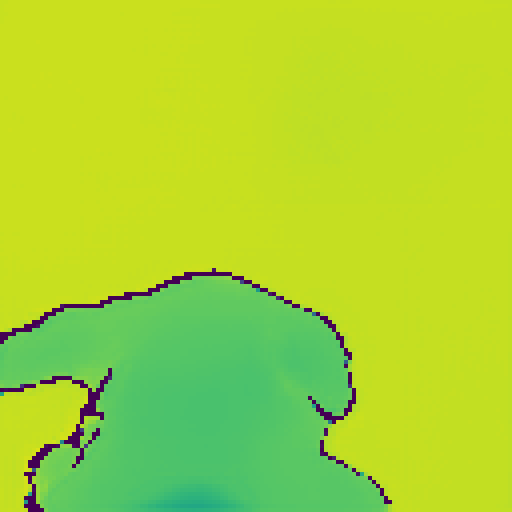}}
	\hfill
	\subfigure{\includegraphics[width=\imgwidth]{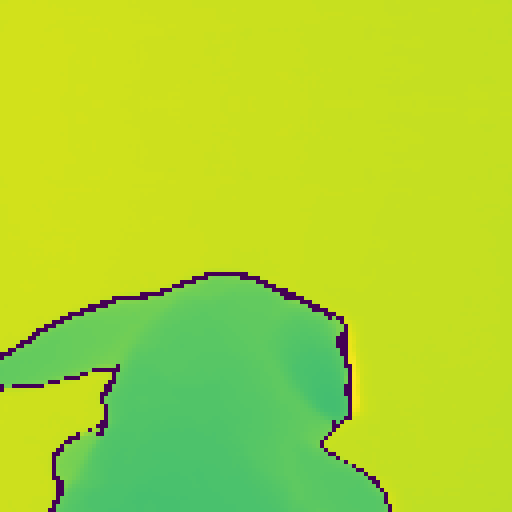}}
	
	\vspace{\largespace}
	
	\renewcommand{\fignum}{12}
	\subfigure{\stackunder[4pt]{\includegraphics[width=\imgwidth]{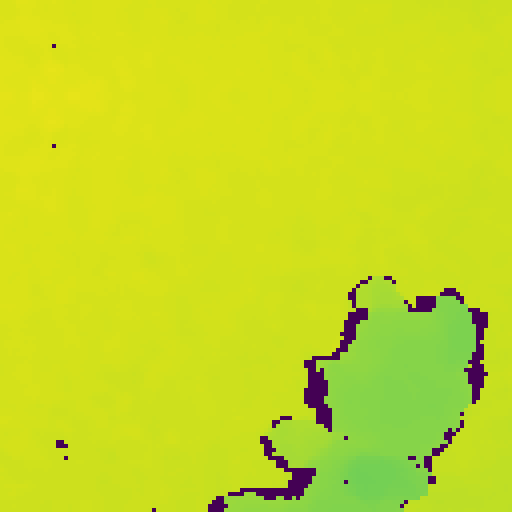}}{\small Input}}
	\hfill
	\subfigure{\stackunder[4pt]{\includegraphics[width=\imgwidth]{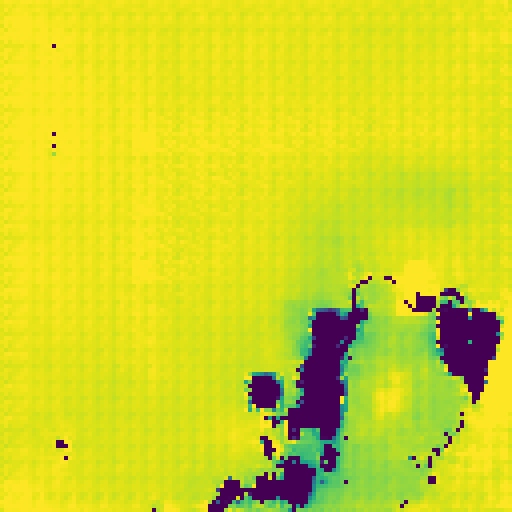}}{\small Cyc. GAN}}
	\hfill
	\subfigure{\stackunder[4pt]{\includegraphics[width=\imgwidth]{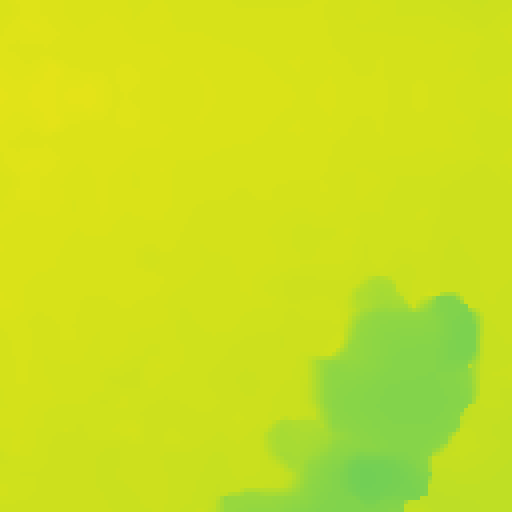}}{\small SDS~\cite{sparse-depth-sensing}}}
	\hfill
	\subfigure{\stackunder[4pt]{\includegraphics[width=\imgwidth]{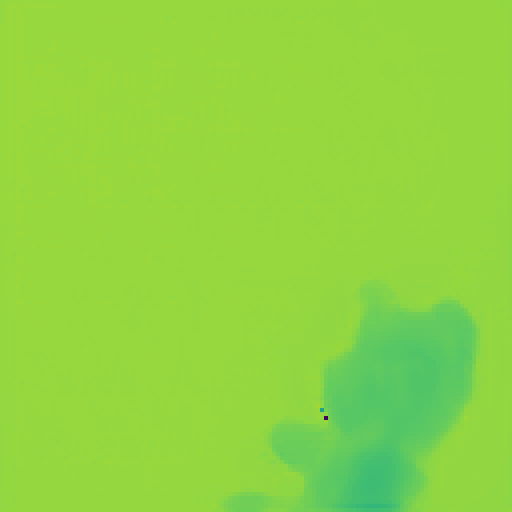}}{\small \textbf{Ours}}}
	\hfill
	\subfigure{\stackunder[4pt]{\includegraphics[width=\imgwidth]{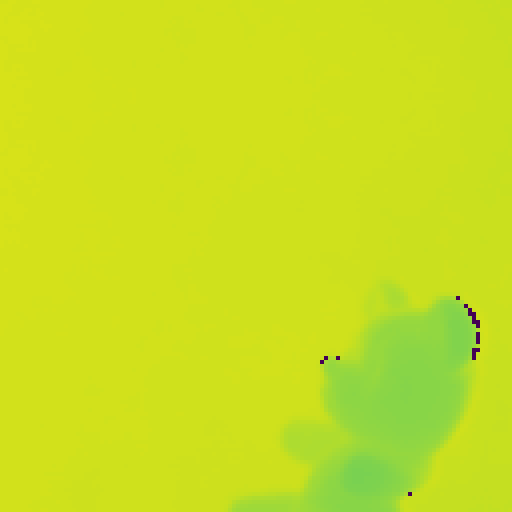}}{\small Kinect 2}}
	
	\caption{Example results for the DROT dataset.}
	\label{fig:results-daniel}
\end{figure}

Table~(\ref{tbl:drot-results-16}) details our quantitative results on this dataset, and Figure~(\ref{fig:results-daniel}) shows some example results.
As can be seen, the original Cycle-GAN remains unusable in this case. However, both our method and Sparse Depth Sensing produce very competitive results, with each exhibiting different visual strengths and artifacts. Specifically, our method produces sharper edges and more accurate geometries and boundaries, whereas~\cite{sparse-depth-sensing} produces images with no missing pixels and with negligible depth shift. 
Perhaps ironically, the main limitation of our method may be its own success --- specifically, as our network was trained to produce convincing \emph{Kinect~2} images, it has also learned to reproduce its typical artifacts and noise patterns, such as missing pixels in this case. Nonetheless, the visual results as well as the higher PNCC scores of our method support its improved reconstruction of geometry and detail in this case.

\begin{table}
	\renewcommand\arraystretch{1}
	\setlength{\tabcolsep}{17pt}
	\centering
	\begin{tabular}{|c|c|c|c|}
		\hline
		\textbf{Base} & \textbf{Tri-Cycle} & \textbf{Sparse Sensing} \\
		\hline
		0.213 & \textbf{0.633} & 0.611 \\
		\hline
	\end{tabular}
	\caption{Quantitative results for the DROT dataset in terms of PNCC. Left to right: base Cycle-GAN, our result, result of~\cite{sparse-depth-sensing}.}
	\label{tbl:drot-results-16}
\end{table}

\section{Conclusions}
\label{sec:conclusion}

Enhancing depth images with real-world noise is an immensely challenging task, with few practical solutions at this point. Formulating the problem as an unsupervised translation task dramatically simplifies dataset construction, however, the existing Cycle-GAN framework is found to be insufficient for this complex task. To overcome this, we proposed several modifications to the framework: a much larger generator architecture designed to handle low-quality depth, use of depth-specific masked similarity losses, and importantly, the asymmetric Tri-Cycle loss which promotes information-preservation between non-equivalent domains. We have tested these modifications on three datasets, and found them to dramatically improve over the base Cycle-GAN in all cases, producing sharp, detailed, and realistic-looking images. We conclude that the proposed approach enables effective enhancement of real-world depth images with severe noise and degradations, expanding the applicability of the Cycle-GAN to asymmetric tasks which do not necessarily satisfy the cycle-consistency assumption.

{\small
\bibliographystyle{ieeetran}
\bibliography{bibliography}
}

\end{document}